\documentclass[runningheads]{llncs}

\usepackage{eccv}

\usepackage{eccvabbrv}

\usepackage{graphicx}
\usepackage{booktabs}

\usepackage[accsupp]{axessibility}  

\usepackage{hyperref}

\usepackage{orcidlink}

\usepackage{amsfonts}
\usepackage{algorithmic}
\usepackage{algorithm}
\usepackage{array}
\usepackage{textcomp}
\usepackage{stfloats}
\usepackage{url}
\usepackage{verbatim}
\usepackage{graphicx}
\usepackage{amsmath}
\usepackage{amssymb}
\usepackage{tabu} 
\usepackage{booktabs} 
\usepackage{lipsum}
\usepackage{mwe}
\usepackage{multirow}
\usepackage{pifont}
\usepackage{tabularx}
\usepackage{colortbl}
\usepackage{fontawesome}
\usepackage{subcaption}
\usepackage{marvosym}

\begin{document}

\title{HiReFF: High-Resolution Feedforward Human Reconstruction from Uncalibrated Sparse-View Video}

\titlerunning{HiReFF: High-Resolution Feedforward Human Reconstruction from Video}

\author{Yiming Jiang\inst{1}\thanks{Y.Jiang—Work done during an internship at Tsinghua University.}\orcidlink{0009-0002-8071-2894} \and
Hanzhang Tu\inst{2}\orcidlink{0009-0003-7555-9546} \and
Wenfeng Song\inst{3}\orcidlink{0000-0002-5101-1071} \and 
Siyou Lin\inst{2}\orcidlink{0000-0002-8906-657X} \and 
Liang An\inst{2}\orcidlink{0000-0002-1028-3759} \and 
Shuai Li\inst{1,4}\orcidlink{0000-0003-4182-1588} \and 
Aimin Hao\inst{1}\textsuperscript{(\Letter)}\orcidlink{0000-0002-5774-6706} \and 
Yebin Liu\inst{2}\orcidlink{0000-0003-3215-0225}
}

\authorrunning{Y.~Jiang, H.~Tu et al.}

\institute{State Key Laboratory of Virtual Reality Technology and Systems, Beihang University, Beijing, China \\
\email{\{jiangyimingjym, lishuai, ham\}@buaa.edu.cn}\\
\and
Tsinghua University, Beijing, China \\
\email{\{thz22, linsy21\}@mails.tsinghua.edu.cn} \\
\email{\{anliang, liuyebin\}@mail.tsinghua.edu.cn}\\
\and
College of Computer Science, Beijing Information Science and Technology University, Beijing, China \\
\email{songwenfenga@gmail.com}\\
\and
Zhongguancun Laboratory, Beijing, China}

\maketitle

\begin{abstract}
Uncalibrated volumetric video streaming for human reconstruction is essential for holographic communication and AR/VR, yet remains challenging due to the need for temporal consistency and computational efficiency from sparse-view inputs. Existing methods rely on per-scene optimization or calibrated cameras, while recent feed-forward models are limited to low-resolution (0.5K) single-frame synthesis.
We present HiReFF, a feed-forward method for 2K-resolution 360° human video reconstruction from uncalibrated sparse-view videos. Our framework decomposes the problem into two key tasks: foreground 3D Gaussian reconstruction from sparse-view videos (four views separated by 90°) and computationally efficient high-resolution synthesis. To enable the former, we propose Scale-synchronized Camera Calibration to resolve scale ambiguity for multi-view supervision, and Gaussian-wise Foreground Masking to reconstruct clean foregrounds by modulating Gaussian parameters. For efficient high-resolution synthesis, our High-resolution Side-tuning achieves 2K rendering by augmenting the Gaussian head with supplementary features while keeping the backbone at 0.5K, drastically reducing computational overhead. Experiments demonstrate that HiReFF significantly outperforms existing methods in high-resolution streaming volumetric video reconstruction. \url{https://iridescentjiang.github.io/HiReFF}
  \keywords{3D reconstruction \and Feedforward 3D gaussian \and Digital human}
\end{abstract}

\begin{figure}
    \centering
    \includegraphics[width=1\textwidth]{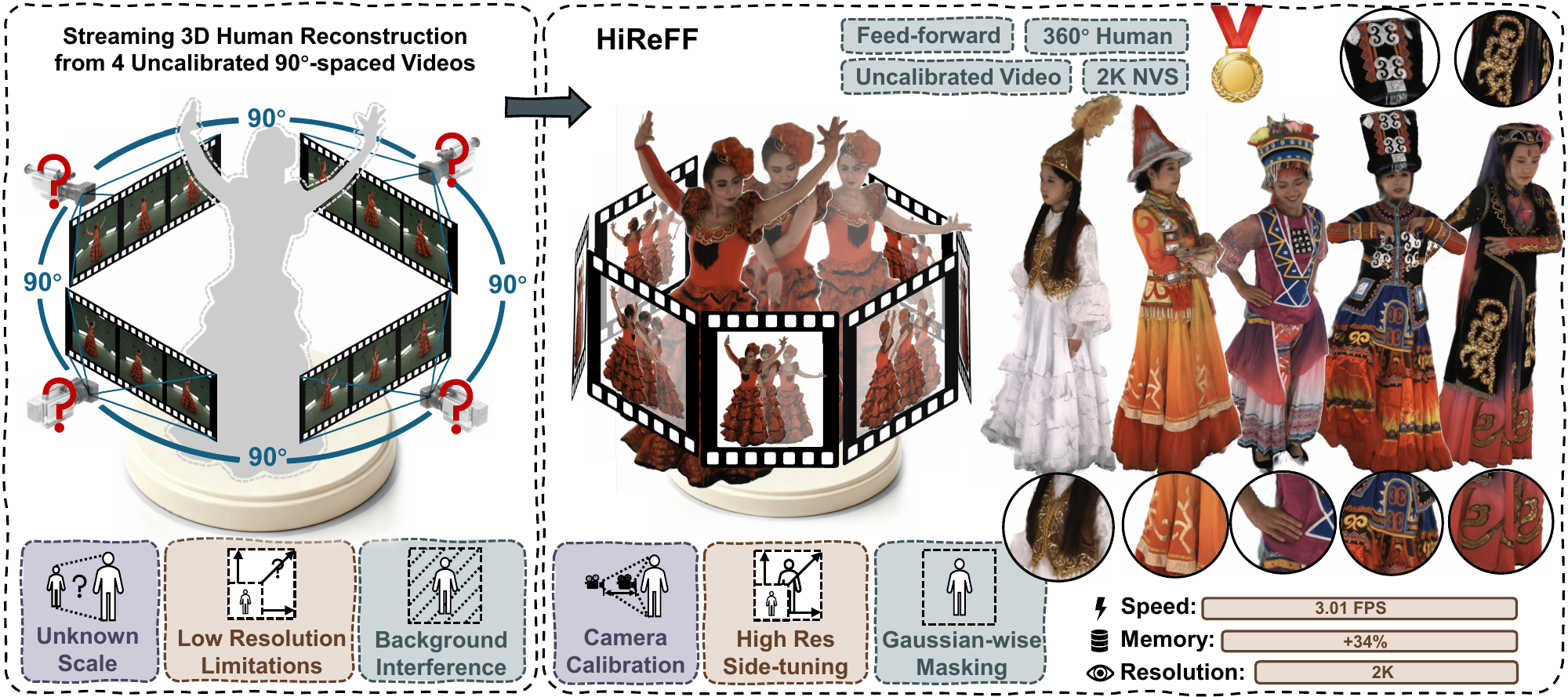}
    \caption{\label{fig:teaser}
        We present HiReFF, a feed-forward method for 2K-resolution 360° human video reconstruction from uncalibrated sparse-view videos. With four-view uncalibrated videos as input, HiReFF reconstructs a 360° human in a streaming fashion at 3.01 FPS on a single RTX 4090 GPU and achieves 2K resolution with only 34\% additional VRAM during training compared to 0.5K.
    }
    \vspace{-0.6cm}
\end{figure}

\section{Introduction}
\label{sec:intro}
Uncalibrated volumetric video streaming for human reconstruction holds diverse applications, including holographic communication~\cite{Tele-Aloha}, augmented/virtual reality (AR/VR)~\cite{SIFU, 3D_Facial, EfficientHuman, Avat3r, DynAvatar, Human4DiT}, and sports broadcasting~\cite{Human_Sports}. Despite recent breakthroughs in static human reconstruction~\cite{Gbc-splat, HumanSplat, hu2025eva, Magicman, PSHuman, HFHuman, DECON, HumanPro}, uncalibrated volumetric video streaming remains exceptionally challenging, as existing methods are either reliant on per-scene iterative optimization over video sequences~\cite{Longvolcap, CAT4D, Lyra, FlashWorld} or necessitate calibrated camera setups~\cite{DoubleField, GPS-Gaussian, Diffuman4D, VolSplat}. The advent of uncalibrated 3D feed-forward reconstruction models~\cite{VGGT, Pi3, MapAnything, StreamVGGT, FastVGGT, 4DNeX} has laid a foundation for addressing this task, demonstrating the ability to predict depth, point clouds, and camera parameters from sparse-view uncalibrated imagery. Concurrently, AnySplat~\cite{AnySplat} has extended these capabilities to achieve photorealistic novel view synthesis from such inputs. Nevertheless, these methods are primarily designed for single-frame inference, present challenges in sparse-view scenarios with wide baselines (e.g., 90° spacing), and are computationally constrained to $518\times518$ resolution outputs. To overcome these limitations, as shown in Fig.~\ref{fig:teaser}, we present HiReFF, a feed-forward method for high-resolution 360° human video reconstruction that enables streaming reconstruction of human models and photorealistic rendering from uncalibrated sparse-view videos.

Our approach enables \textbf{(1) the reconstruction of foreground human 3D Gaussians from uncalibrated, streaming sparse-view video inputs} (four views separated by 90°), and \textbf{(2) efficient high-resolution novel view synthesis without incurring substantial computational overhead.} 
To achieve the first objective, we leverage pretrained parameters from the feed-forward 3D reconstruction model VGGT~\cite{VGGT} to obtain predicted camera parameters and depth priors, and augment it with Gaussian and mask prediction heads to reconstruct human 3D Gaussians. In doing so, we address two key challenges:
1) The scale mismatch hindering additional view supervision. Our setup involves four 90°-spaced input views, rendering supervision from only these views is insufficient. We therefore require additional novel viewpoint supervision, which necessitates ground-truth camera parameters. This presents a fundamental challenge: VGGT's predictions lack metric scale, preventing direct alignment with the ground-truth camera. To address this, we propose Scale-synchronized Camera Calibration (Sec.~\ref{sec:Method Pipeline}), which dynamically adjusts camera parameters for extra supervision views during training and employs indirect supervision of the camera head.
2) Foreground reconstruction compromises camera estimation. Directly masking input images for foreground-only reconstruction severely degrades camera parameter accuracy in widely-spaced (90°) sparse-view settings, creating a conflict between foreground masking and camera prediction. Therefore, we propose Gaussian-wise Foreground Masking (Sec.~\ref{sec:Method Pipeline}), which introduces a mask head to modulate Gaussian parameters. Approximately orthogonal input views may leave a small number of extraneous Gaussians after masking; however, our experiments show that they vanish during training, yielding a clean foreground reconstruction.

To achieve computationally efficient high-resolution reconstruction, we propose High-resolution Side-tuning (Sec.~\ref{sec:Method High-resolution}), inspired by side-tuning~\cite{Side-Tuning}, which augments the Gaussian head with supplementary features to enable high-resolution output. Constrained by computational overhead, prior models~\cite{AnySplat} are limited to $518\times518$ novel-view rendering, which proves inadequate for practical high-resolution applications. Critically, however, high-resolution rendering does not necessitate higher precision in VGGT features, but rather demands enhanced appearance attributes. Consequently, we only increase the number of depth and Gaussian positions through interpolation and employ a supplementary architecture that injects additional image-derived features into the mid-level of the Gaussian head, thereby maintaining the backbone's input resolution at $518\times518$.
During training, we render novel-view images at high resolution and supervise against high-resolution ground-truth imagery, while employing patch-wise perceptual loss to further mitigate computational burden. Our high-resolution rendering results, together with measurements of memory footprint and computation time, substantiate the effectiveness of this approach.

The contributions of this work are summarized as follows:

\begin{itemize}
\item We introduce HiReFF, a feed-forward framework for high-resolution 360° human video reconstruction, achieving 2K-resolution streaming 3D Gaussian reconstruction from four uncalibrated 90°-spaced views.
\item We propose Scale-synchronized Camera Calibration to resolve metric scale ambiguity for effective multi-view supervision, and Gaussian-wise Foreground Masking to reconstruct clean foregrounds by modulating Gaussian parameters.
\item We design High-resolution Side-tuning that achieves efficient 2K rendering by augmenting the Gaussian head with supplementary image features while maintaining the backbone at 518×518 resolution, significantly reducing computational overhead.
\item Extensive experiments on benchmark datasets demonstrate that our method significantly outperforms existing approaches in high-resolution streaming volumetric video reconstruction, both quantitatively and qualitatively.
\end{itemize}

\section{Related Work}
\label{sec:Related Work}

\textbf{Feed Forward Reconstruction.} Significant progress has been made in feed-forward 3D reconstruction from multi-view images. To achieve photorealistic novel view synthesis, a line of research~\cite{MVSplat, GGN, FreeSplat, DepthSplat, HGM, Gbc-splat, Splat-SAP, lin2025depth} focuses on directly predicting static 3D Gaussians from calibrated multi-view images. Meanwhile, another line of methods, such as DUSt3R~\cite{DUSt3R}, FLARE~\cite{FLARE}, VGGT~\cite{VGGT}, and MapAnything~\cite{MapAnything} can now directly regress camera poses and 3D point clouds within the coordinate frame of the first image, while $\pi$3~\cite{Pi3} extends this capability to relative coordinate systems. Furthermore, to reduce reliance on camera calibration, methods such as NoPosplat~\cite{NopoSplat} and AnySplat~\cite{AnySplat} aim to reconstruct 3D Gaussians from uncalibrated images. Recently, leveraging the capabilities of uncalibrated 3D reconstruction models, FastAvatar~\cite{FastAvatar} and Human3R~\cite{Human3R} have extended 3D Gaussian reconstruction from uncalibrated images to human reconstruction. However, the high computational cost of geometry transformers limits feed-forward 3D reconstruction networks to input resolutions around 0.5K, preventing direct high-resolution novel view synthesis. To address this, our method is designed for efficient high-resolution rendering. It can quickly reconstruct high-quality 3D Gaussian and render 2K images without a significant computational penalty.

\textbf{Sparse-view Human Reconstruction.} 
Reconstructing dynamic humans from sparse camera arrays is crucial for practical deployment, yet severely ill-posed due to limited geometric cues. 
While some approaches attempt reconstruction via parametric avatars~\cite{Vid2Actor, AniGS, AnimatableGaussians, CtrlAvatar}. IDOL~\cite{Idol} and LHM~\cite{LHM} can reconstruct an avatar in several seconds using human-specific priors such as SMPL~\cite{SMPL}. However, they encounter fundamental difficulties in modeling loose clothing and complex deformations.
Other methods~\cite{HumanNeRF, 3dgs-avatar} rely on per-scene optimization, demanding calibrated multi-view rigs and prohibitive training times. To relax these constraints, feed-forward methods~\cite{DoubleField, DiffuStereo, GPS-Gaussian, RoGSplat, Diffuman4D} learn generalizable priors from large-scale datasets, enabling faster reconstruction but still requiring precise camera calibration.
Recently, Forge4D~\cite{Forg4D} first achieved frontal human novel-view synthesis from uncalibrated videos.
Consequently, no existing method simultaneously addresses the core challenges of uncalibrated sparse-view inputs, streaming video reconstruction, and high-resolution 360° rendering—all critical for real-world applications in holographic communication and immersive broadcasting. HiReFF closes this gap by introducing a feed-forward framework that achieves 3D Gaussian reconstruction at 2K resolution without camera calibration.

\textbf{High-Resolution Novel View Synthesis.}
High-resolution Novel View Synthesis (HRNVS) targets high-resolution novel view generation from low-resolution (LR) inputs. Early NeRF-based methods~\cite{NeRF-SR, RefSR-NeRF, Super-NeRF} pioneered optimizing high-resolution neural radiance fields under sub-pixel constraints. Recently, 3D Gaussian Splatting (3DGS) capitalizes on its advantage of faster rendering speeds to produce high-quality imagery~\cite{deng2024compact, Deng_2026_CVPR}. Subsequent render methods~\cite{Mip-Splatting, Arbitrary-ScaleGS} have been proposed to address the rendering clarity of 3D Gaussian across resolutions. Super-resolution for 3D Gaussian Splatting has been addressed in several recent works. SRGS~\cite{SRGS} achieves high-quality HRNVS using only existing low-resolution views, while GaussianSR~\cite{GaussianSR} incorporates 2D diffusion priors for super-resolution. However, these methods are not readily compatible with modern feed-forward 3D reconstruction backbones. To bridge this gap, we introduce a sidetune adaptation strategy that enables fast and computationally efficient HRNVS within a feed-forward 3DGS framework.

\vspace{-6pt}
\begin{figure*}[h!]
	\begin{center}
		\includegraphics[clip, width=\linewidth]{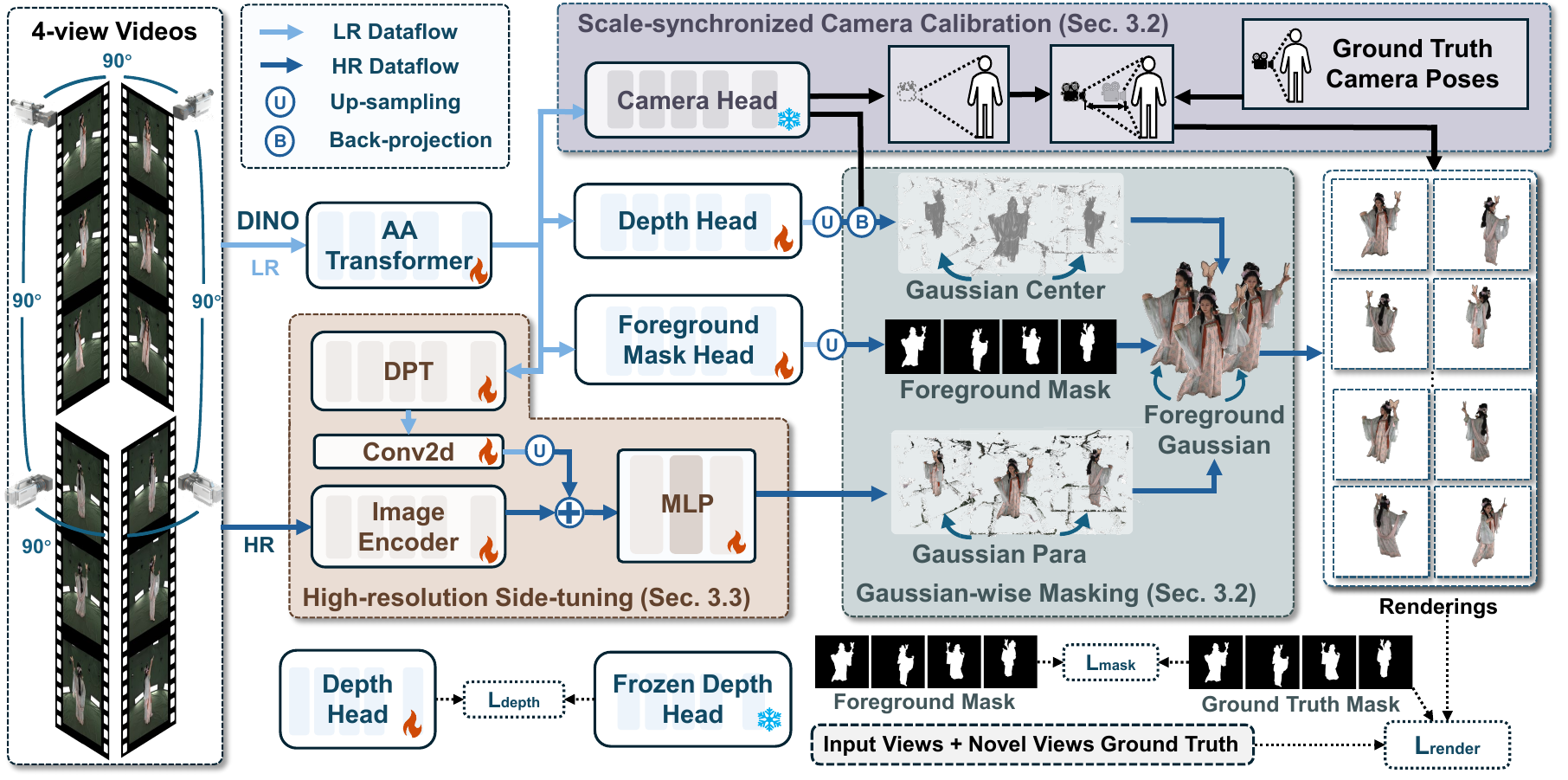}
		\caption{~\textbf{Method Overview} (\S \ref{sec:Method}). Taking four-view uncalibrated videos as input, we first extract features using an Alternating-Attention (AA) Transformer, then decode to obtain Gaussian parameters, supervising through rendered multi-view images. Specifically, HiReFF employs Scale-synchronized Camera Calibration (Sec.~\ref{sec:Method Pipeline}) to introduce supervision from additional viewpoints while indirectly supervising the Camera Head, uses Gaussian-wise Masking (Sec.~\ref{sec:Method Pipeline}) to remove background while preserving camera parameter accuracy, and leverages High-resolution Side-tuning (Sec.~\ref{sec:Method High-resolution}) to ensure computational efficiency at high resolution.}
		\label{fig: pipeline}
	\end{center}
    \vspace{-1.2cm}
\end{figure*}

\section{Method}
\label{sec:Method}

As shown in Fig.~\ref{fig: pipeline}, we present a method that achieves 2K-resolution feed-forward human volumetric video reconstruction: from just four uncalibrated videos captured 90° apart, our network rapidly reconstructs the 3D Gaussian capable of being rendered at 2K resolution from any novel 360° viewpoint.

The section is structured as follows: the problem is formulated in Sec~\ref{sec:Method Problem Setup}, the overall architecture and pipeline are introduced in Sec~\ref{sec:Method Pipeline}, and our technique for computationally efficient high-resolution reconstruction is detailed in Sec~\ref{sec:Method High-resolution}.

\subsection{Problem Setup}
\label{sec:Method Problem Setup}
Consider four uncalibrated high-resolution (HR) RGB videos capturing a dynamic scene from viewpoints spaced 90° apart, denoted as \( \{ V_i \}_{i=1}^4 \), where each video \( V_i \) consists of frames \( I_i^t \in \mathbb{R}^{H_\text{HR} \times W_\text{HR} \times 3} \). Our goal is to reconstruct a HR 3D Gaussian Splatting (3DGS) volumetric video by estimating, for each time step \( t \):
1. A set of \( G \) anisotropic 3D Gaussians
$\{ (\mu_g^t, \sigma_g^t, r_g^t, s_g^t, c_g^t)\}_{g=1}^G,$
where \( G = 4 \times H_\text{HR} \times W_\text{HR} \) corresponds to the total number of pixels across all four input HR views, parameterized by position \( \mu \in \mathbb{R}^3 \), opacity \( \sigma \in \mathbb{R}^+ \), rotation \( r \in \mathbb{R}^4 \), scale \( s \in \mathbb{R}^3 \), and spherical harmonic coefficients \( c \in \mathbb{R}^{3 \times (k+1)^2} \) of degree \( k \);
2. Temporally smooth per-view camera parameters \( \{ p_i^t \in \mathbb{R}^9 \}_{i=1}^4 \) for each view, estimated with inter-frame continuity constraints to ensure stable trajectory reconstruction.
Formally, our model performs the mapping from high-resolution multi-view videos to a high-resolution dynamic 3DGS representation:
$f_\theta: \{ V_i^\text{HR} \}_{i=1}^4 \longmapsto \left\{ ( \mu_g^t, \Theta_g^t )_{g=1}^G \cup ( p_i^t )_{i=1}^4 \right\}_{t=1}^T,$
enabling high-resolution novel view synthesis at \( H_\text{HR} \times W_\text{HR} \) resolution and arbitrary temporal intervals. \( \Theta_g^t = ( \sigma_g^t, r_g^t, s_g^t, c_g^t ) \) represents the 3DGS parameters.

The model is evaluated on temporally consistent 3D reconstruction and high-resolution novel view synthesis at \( H_\text{HR} \times W_\text{HR} \), and also produces auxiliary outputs such as per-frame depth maps and smooth camera trajectories usable in volumetric video applications.

\subsection{Inter-frame Scale-stabilized Feedforward 3D Gaussian Reconstruction}
\label{sec:Method Pipeline}
Recent foundational models for uncalibrated feedforward 3D reconstruction~\cite{VGGT, MapAnything} employ an Alternating-Attention (AA) Transformer to extract features, which are subsequently decoded into basic 3D information (camera parameters, depth, point clouds), yet they fail to produce photorealistic imagery. 

\textbf{Side-tuning Gaussian Prediction.} Recently, AnySplat~\cite{AnySplat} demonstrated the feasibility of feedforward photorealistic synthesis by utilizing a DPT~\cite{DPT} as a Gaussian head $F_G$ to decode AA Transformer features into 3D Gaussian Splatting (3DGS) parameters. Inspired by AnySplat, we introduce a side-tuning~\cite{Side-Tuning} approach where features extracted from images via an MLP $F_a$ are combined with intermediate $F_G$ features to predict per-frame 3DGS parameters for videos. Specifically, let \( I^t \in \mathbb{R}^{H \times W \times 3} \) denote the video frame at time \( t \). The 3DGS parameters $\Theta^t$ are obtained through:
\vspace{-2pt}
\begin{equation}
\begin{split}
& f_{A}^t = F_{AA}(I^t), \\
& \Theta^t = F_D\big(\,F_a(I^t) \oplus F_G^{\text{mid}}(f_{A}^t)\,\big),
\end{split}
\end{equation}
where \(F_{AA}\) represents AA Transformer, $f_{AA}$ represents the feature encoded by \(F_{AA}\), \( F_G^{\text{mid}}(\cdot) \) represents intermediate features from the Gaussian head, \( \oplus \) indicates adding operation, \( F_D(\cdot) \) is another MLP generates the final 3DGS parameters including Gaussian attributes \( \{ \sigma, r, s, c\} \). Position $\mu$ is obtained via back-projection using depth predictions from the depth head and camera parameters from the camera head, thereby establishing per-pixel alignment with the input image. Specifically, for a pixel with homogeneous coordinates $\mathbf{p} = [u, v, 1]^T$, predicted depth $d$, camera intrinsics $\mathbf{K}$, and extrinsics $[\mathbf{R} \,|\, \mathbf{t}]$, the corresponding 3D point in world coordinates is: $\mu = \mathbf{R}^T \left( d \, \mathbf{K}^{-1} \mathbf{p} - \mathbf{t} \right),$
where $R$ is the $3 \times 3$ rotation matrix and $\mathbf{t}$ is the $3$-dimensional translation vector of the camera pose.

The proposed sidetune module augments the Gaussian DPT head by providing crucial low-level, fine-grained details from the image, serving as a crucial complement to the high-level features encoded by the AA Transformer.


\textbf{Gaussian-wise Foreground Masking.} For human reconstruction tasks, the objective is to reconstruct only the foreground human region while masking out the surrounding background. However, with four input views spaced 90° apart exhibiting minimal overlap, using only the foreground human regions presents significant challenges for camera parameter estimation. We empirically observe that larger input regions yield more accurate camera predictions. To address this, we employ the full original images as input and introduce a mask head $F_m$ to selectively filter the predicted Gaussians, ensuring that only the foreground human region is reconstructed. Formally, this process is defined as:
\vspace{-2pt}
\begin{equation}
\begin{split}
(\Theta^t_{g}, \mu_g^t)_{g=1}^{\text{H}} = F_m((\Theta^t_{g}, \mu_g^t)_{g=1}^G) \odot ( \Theta^t_{g}, \mu_g^t)_{g=1}^G,
\end{split}
\end{equation}
where \(\odot\) denotes Gaussian-wise masking based on the predicted foreground probabilities. $(\Theta^t_{g}, \mu^t_{g})_{g=1}^{\text{H}}$ denotes the Gaussian parameters constituting the human subject to be reconstructed.


\textbf{Scale-Synchronized Camera Calibration.} Previous methods~\cite{AnySplat} for supervising the camera head typically employ direct supervision using ground-truth camera parameters, while simultaneously rendering 3D Gaussians based on the camera head's predictions and supervising these renderings with real images. However, we observe that for inputs with large viewpoint intervals (e.g., 90°), supervision from only the input views proves insufficient, necessitating the introduction of additional supervision viewpoints. Moreover, such large viewpoint intervals cause the camera head to exhibit significant fluctuations during training. These fluctuations substantially impact the final rendering outcome, causing a large variance in the rendering loss, which hinders convergence.
To address this, we freeze the camera head during training and instead use the ground-truth camera parameters from both the input and supervision viewpoints to render the Gaussian parameters \(\mu^t\) and \(\Theta^t\), while activating the AA Transformer to modify the input features to the camera head, thereby indirectly optimizing it. This approach introduces novel viewpoints for supervision and avoids large fluctuations in the camera head, leading to improved convergence.
Since the base model lacks a true scale—the initialized positions \(\mu^t\) of the Gaussian points only align with the predicted camera parameters and do not possess a realistic scale (e.g., the predicted human body might be only 0.8m tall)—it is infeasible to directly use the true camera parameters to render \(\mu^t\). Therefore, we scale the true camera parameters by adjusting the translation \(\hat{\mathbf{t}}\) to match the scale of the predicted translation \(\mathbf{t}\). 
Furthermore, we observe that this indirect supervision approach maintains the stability of the camera head’s predictions across frames, effectively mitigating scale fluctuation in inter-frame reconstruction results.

Specifically, for each non-reference view \(i\) (where \(i = 2, 3, \dots, V\), with \(V\) being the number of views, and \(i=1\) as the reference view), the scale adjustment is computed as:
\vspace{-8pt}
\begin{equation}
\begin{split}
& \bar{s} = \frac{1}{V-1} \sum_{i=2}^{V} \frac{\hat{\mathbf{t}}_i }{ \mathbf{t}_i }, \\
& \hat{\mathbf{t}}'_i = \frac{\hat{\mathbf{t}}_i}{\bar{s}} \quad \text{for } i = 2, \dots, V.
\end{split}
\end{equation}
\vspace{-2pt}
Subsequently, we use the true intrinsic parameters \(\hat{\mathbf{K^t}}\), rotation matrices \(\hat{\mathbf{R^t}}\), and the adjusted translation \(\hat{\mathbf{t^t}}'\) to render the Gaussian parameters. While the position $\mu$ is back-projection using the predicted camera parameter $R^t$, $K^t$, and $t^t$. The rendering process can be formulated as:
\vspace{-2pt}
\begin{equation}
\begin{split}
& \mu_g^t = \mathbf{R^t}^T \left( d^t \, \mathbf{K^t}^{-1} \mathbf{p} - \mathbf{t^t} \right), \\
& I^t = \mathcal{R}( \{\mu_g^t, \Theta_g^t\}_{g=1}^{H}, \hat{\mathbf{K^t}}, \hat{\mathbf{R^t}}, \hat{\mathbf{t^t}}' ),
\end{split}
\end{equation}
where \(\mathcal{R}(\cdot)\) denotes the rendering function that projects the 3D Gaussians onto the image plane using the given camera parameters, producing the rendered image \(I^t\). \(I^t\) is then used for supervision with the ground truth image \(\hat{I^t}\).

\subsection{Computationally Efficient High-resolution Reconstruction}
\label{sec:Method High-resolution}
Volumetric video reconstruction demands high-resolution rendering. However, feeding high-resolution images (e.g., 2K) directly into the AA Transformer of feed-forward 3D foundation models incurs excessive computational cost—existing models~\cite{VGGT, AnySplat} typically operate at $518\times518$ resolution. Thus, we need to seek an approach that enhances reconstruction resolution while preserving the AA Transformer input resolution.

~\textbf{High-resolution Side-tuning.} Importantly, high-resolution rendering quality depends primarily on the high-dimensional Gaussian attributes \(\Theta_g\) output by the Gaussian head \(F_G\), rather than on position \(\mu_g\) or camera position \(p_i\). Therefore, we maintain the input of AA Transformer at $518\times518$ while providing complementary high-resolution information to \(F_G\) through a supplementary pathway.
Specifically, we employ a side-tuning strategy: high-resolution images \(I_{HR}\) are fed to the supplementary network \(F_a\) while low-resolution images \(I_{LR}\) are processed by the AA Transformer. Intermediate features \(F_G^{\text{mid}}(I_{LR})\) are upsampled to high resolution before fusion with \(F_a(I_{HR})\) and subsequent feeding into \(F_D\), formulated as:
\vspace{-2pt}
\begin{equation}
\Theta^t = F_D\Big(F_a\big(I_{HR}^t\big) \oplus \mathcal{U}\big(F_G^{\text{mid}}(I_{LR}^t)\big)\Big),
\end{equation}
where \(\mathcal{U}(\cdot)\) denotes upsampling to high resolution and \(F_a\) is an MLP modified by EdgeNeXt~\cite{Maaz2022EdgeNeXt}.
Subsequently, the mask head \(F_m\) and depth head \(F_d\) outputs are upsampled to high resolution, and rendering is performed at high resolution for supervision against ground-truth high-resolution images.


\textbf{Training Loss.}
We employ a rendering loss to supervise image synthesis. During training, in addition to the four input views, we sample $V_a$ additional novel viewpoints from frontal, top, and bottom positions outside the input viewing sphere, yielding $4+V_a$ supervision viewpoints in total. We render the network outputs of all $4+V_a$ views and supervise them against ground-truth images. The rendering loss combines L1 loss and perceptual loss, formulated as:
\vspace{-6pt}
\begin{equation}
\mathcal{L}_{\text{render}} = \sum_{i=1}^{4+V_a} \|\hat{I}_i - I_i\|_1 + \lambda_{\text{P}} \sum_{i=1}^{4+V_a} \text{Perceptual}(\hat{I}_i, I_i),
\end{equation}
where \(\hat{I}_i\) and \(I_i\) denote the rendered and ground-truth images for view \(i\), respectively. The perceptual loss~\cite{Perceptual_Loss} measures the distance in the VGG~\cite{VGG}'s feature space, as Animatable Gaussians~\cite{AnimatableGaussians}. We use computation graph surgery~\cite{mescheder2025sharp} to lower the memory consumption of perceptual loss.

For mask supervision, we employ L1 loss between the predicted and ground-truth masks in 4 input views:
\vspace{-8pt}
\begin{equation}
\mathcal{L}_{\text{mask}} = \sum_{i=1}^{4} \|\hat{M}_i - M_i\|_1,
\end{equation}
where \(\hat{M}_i\) and \(M_i\) are the predicted and ground-truth masks for view \(i\).

To mitigate overfitting of the depth head to input viewpoints, we utilize a depth distillation loss. Specifically, we maintain a frozen reference depth head initialized with VGGT~\cite{VGGT} parameters and enforce consistency with our active depth head via MSE loss:
\vspace{-8pt}
\begin{equation}
\mathcal{L}_{\text{depth}} = \sum_{i=1}^{4} \|\hat{D}_i^{\text{a}} - \hat{D}_i^{\text{f}}\|_2^2,
\end{equation}
where \(\hat{D}_i^{\text{a}}\) and \(\hat{D}_i^{\text{f}}\) denote depth predictions from the active and frozen heads for view \(i\), respectively.
The overall training objective combines these losses with weighted coefficients:
\vspace{-4pt}
\begin{equation}
\mathcal{L}_{\text{total}} = \lambda_{\text{render}} \mathcal{L}_{\text{render}} + \lambda_{\text{mask}} \mathcal{L}_{\text{mask}} + \lambda_{\text{depth}} \mathcal{L}_{\text{depth}}.
\end{equation}

\section{Experiment}
\label{sec:Experiment}

\subsection{Experimental Settings}
\label{sec:Experiment Settings}

\textbf{Datasets.}
We conduct training mainly on the DNA-Rendering~\cite{DNA-Rendering} dataset, which encompasses 153 actors and 439 distinct motion sequences. Each sequence comprises 48-view synchronized RGB video streams at $2448\times2048$ resolution, together with corresponding viewpoint camera parameters. To improve generalization, we also utilize the ZJU-MoCap~\cite{peng2021neural,fang2021mirrored} and MVHumanNet~\cite{xiong2024mvhumannet,li2025mvhumannet++} datasets. For validation, we select 20 motion sequences with distinct identities from within the DNA-Rendering dataset as the test set. Detailed dataset configurations and additional results on more datasets are provided in the supplementary materials.

\textbf{Evaluation Metrics.}
We assess novel-view rendering quality using PSNR, SSIM, and LPIPS metrics at $2072\times2072$ resolution. For models rendering at alternative resolutions (e.g., GPS-Gaussian~\cite{GPS-Gaussian}, AnySplat~\cite{AnySplat}), we bilinearly upsample their outputs to 2K. All methods are evaluated under a 4-view input setting with approximately 90° angular separation, providing comprehensive 360° coverage of the human subject. For models requiring camera parameters as input, we supply ground-truth parameters, with this condition clearly indicated in our experimental results.

\textbf{Baselines.}
We compare our approach against existing novel view synthesis methods. Specifically, for uncalibrated camera methods, we benchmark against AnySplat~\cite{AnySplat} and NoPoSplat~\cite{NopoSplat}. We also include comparisons with 4DGT~\cite{4DGT}, a monocular volumetric video reconstruction method. For calibrated camera approaches, we evaluate against GPS-Gaussian~\cite{GPS-Gaussian}.
Since our task involves foreground human segmentation and the comparison methods lack built-in foreground masks, we employ an open-source video portrait segmentation algorithm~\cite{RVM} to generate masks. GPS-Gaussian receives segmented foreground images as input, whereas other methods (including ours) take full images with background intact. To ensure fair comparison, we multiply all methods' predicted results by a consistent mask before computing evaluation metrics.
Please refer to the supplementary materials for experimental setup details.

\textbf{Implementation Details.}
The AA Transformer, camera head, and both active and frozen depth heads are initialized with pretrained weights from VGGT~\cite{VGGT}, whereas the 3D Gaussian prediction head, side-tuning supplementary network, and mask head are zero-initialized. Our Gaussian splatting renderer is based on gsplat~\cite{gsplat}.
We define HR resolution as $2072\times2072$ and LR resolution as $518\times518$. The network receives four HR images from viewpoints separated by 90°. For supervision, we render at HR resolution ($2072\times2072$) and provide ground-truth images at the same scale. 
We set $V_a=4$, supervising across $4+ V_a = 8$ total viewpoints. Loss weights are configured as $\lambda_{\text{P}} = 0.1$, $\lambda_{\text{render}} = 1.0$, $\lambda_{\text{mask}} = 5 \times 10^{-2}$, and $\lambda_{\text{depth}} = 10^{1}$. All training stages are performed on 8 A800 GPUs. 
Automatic mixed precision is employed to accelerate training.

\begin{figure*}[h!]
	\begin{center}
 	\setlength{\abovecaptionskip}{0cm}
		\includegraphics[clip, width=\linewidth]{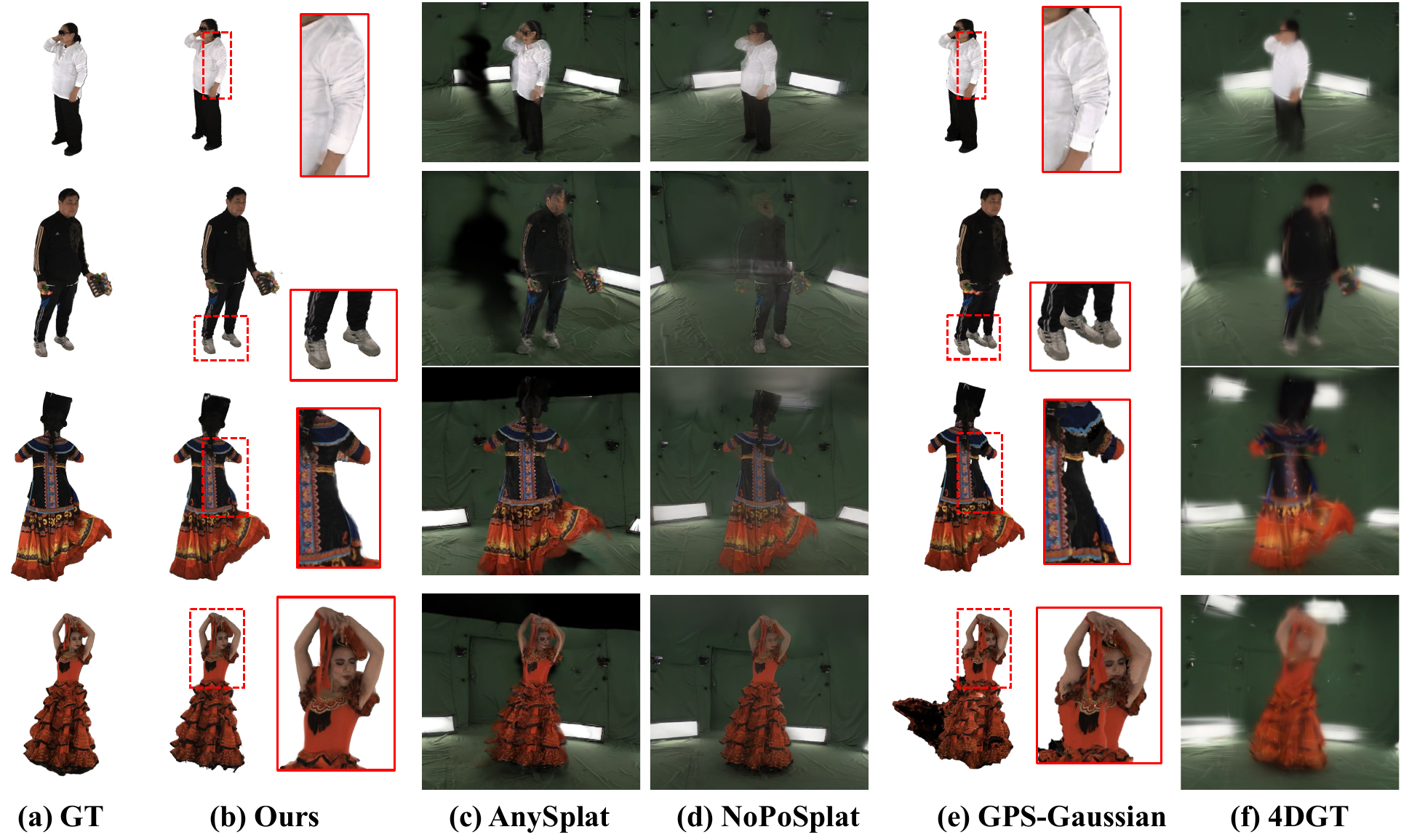}
        \caption{~\textbf{Qualitative results of novel-view synthesis of reconstruction} (\S\ref{sec:Evaluation}). Our method surpasses all others in global shape, garment details, and facial fidelity.}
    \label{fig:qualitative_evaluation}
	\end{center}
\vspace{-0.6cm}
\end{figure*}

\subsection{Evaluation}
\label{sec:Evaluation}

\begin{figure*}[h!]
	\begin{center}
     	\setlength{\abovecaptionskip}{0cm}
		\includegraphics[clip, width=0.95\linewidth]{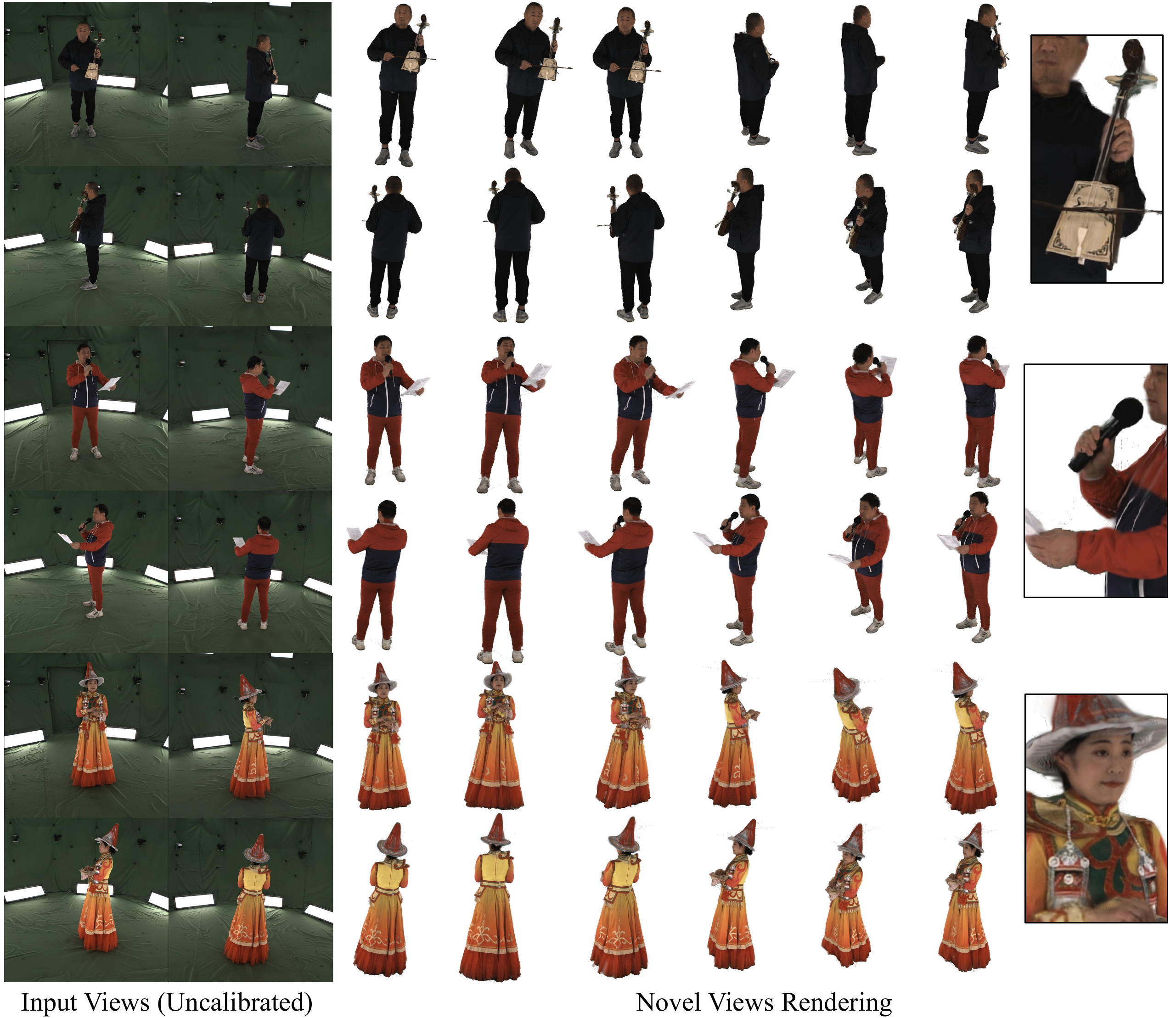}
		\caption{\textbf{More visualization results.} For both top-down and bottom-up perspectives, our method successfully reconstructs the correct geometry and accurately reproduces surface coloration. Please \faSearch ~\textbf{zoom in} to see details.}
		\label{fig: all_views_supmat}
	\end{center}
    \vspace{-0.4cm}
\end{figure*}

\begin{figure*}[h!]
	\begin{center}
     	\setlength{\abovecaptionskip}{0cm}
		\includegraphics[clip, width=0.95\linewidth]{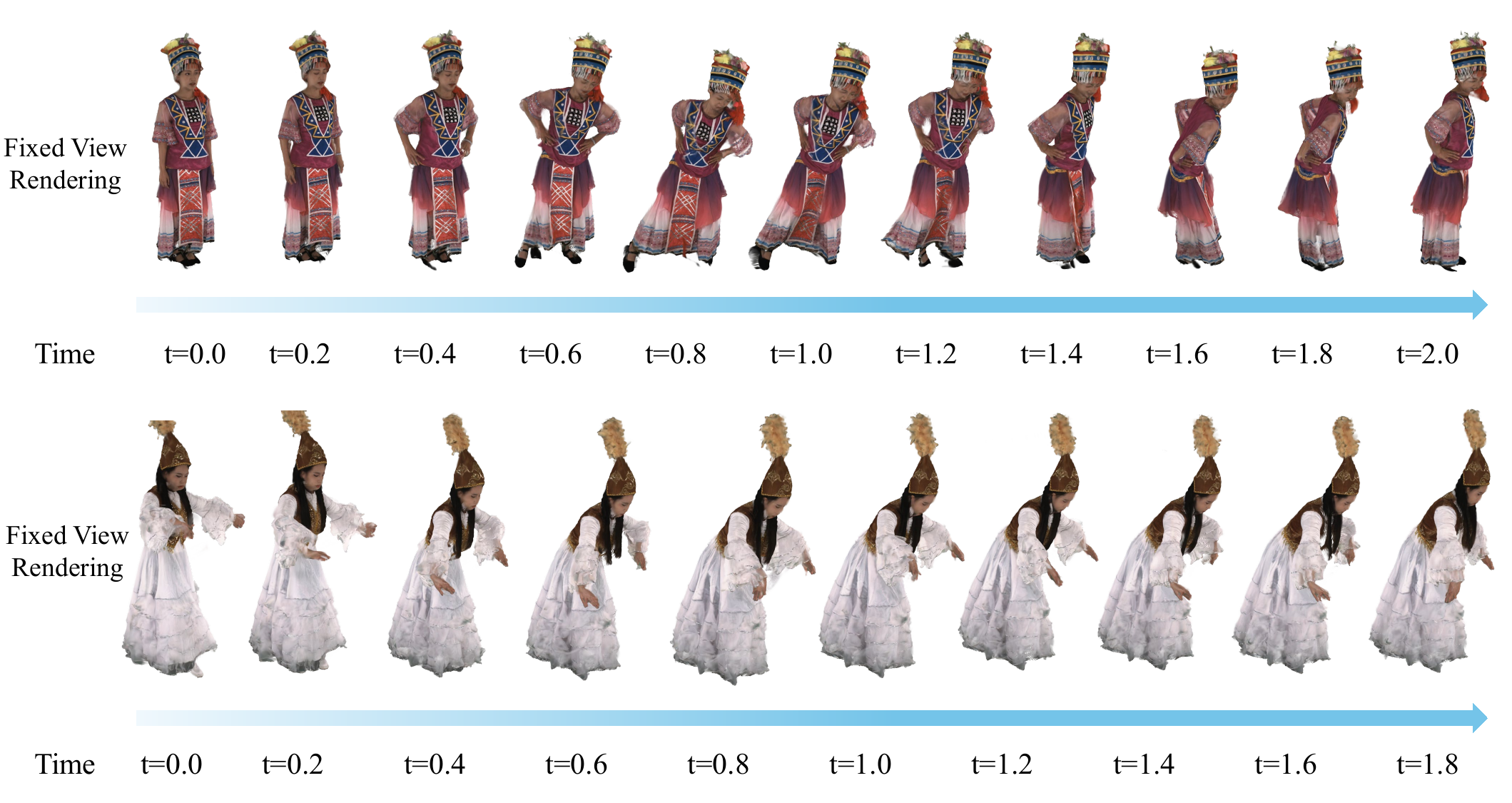}
		\caption{\textbf{Qualitative results on timing consistency.} Our method maintains excellent temporal consistency for both human body and clothing details when reconstructing garments with complex patterns. Please \faSearch ~\textbf{zoom in} to see details.}
		\label{fig:time_consist}
	\end{center}
    \vspace{-0.8cm}
\end{figure*}

\textbf{Qualitative Evaluation.}
As shown in Fig.~\ref{fig:qualitative_evaluation}, our method surpasses all others in global shape, garment details, and facial fidelity. AnySplat~\cite{AnySplat} and NoPoSplat~\cite{NopoSplat} struggle to estimate accurate camera parameters under the large 90° baseline, causing misaligned reconstructions across views. GPS-Gaussian~\cite{GPS-Gaussian}, supplied with eight calibrated views, ground-truth cameras, and pre-segmented foreground images, yields a comparatively complete model, yet still exhibits leg-artifacts. 4DGT~\cite{4DGT} fails to produce plausible results when only a single-frame video is provided. As shown in Fig.~\ref{fig: all_views_supmat}, we present test results from additional novel viewpoints. For both top-down and bottom-up perspectives, our method successfully reconstructs the correct geometry and accurately reproduces surface coloration. In the bottom two rows showcasing complex clothing, our method correctly reconstructs garment patterns. As shown in Fig.~\ref{fig:time_consist}, our method maintains excellent temporal consistency for both human body and clothing details when reconstructing garments with complex patterns. Additional dynamic results are available in our supplementary video. 

\vspace{-6pt}

\begin{table*}[h!]
\centering
\caption{~\textbf{Quantitative results of novel-view synthesis of reconstruction (\S\ref{sec:Evaluation}).} Ours outperforms all uncalibrated camera input methods across all metrics.} 
\renewcommand{\arraystretch}{1.0}
\begin{tabularx}{\linewidth}{@{\hspace{0.0cm}}>{\centering\arraybackslash}p{2.1cm} >{\centering\arraybackslash}p{1.6cm} >
{\centering\arraybackslash}p{1.8cm} >{\centering\arraybackslash}p{1.0cm} >{\centering\arraybackslash}p{1.7cm} >{\centering\arraybackslash}p{1.7cm} >{\centering\arraybackslash}p{1.7cm}}
\toprule
Method & Cam. Pose & Views Num. & Aligned & PSNR↑ & SSIM↑ & LPIPS↓\\
\midrule
GPS-Gaussian & $\checkmark$ & 8 & - & \cellcolor[rgb]{1,0.9,0.6}{26.1039} & \cellcolor[rgb]{1.0,0.7,0.3}{0.9172} & \cellcolor[rgb]{1,0.9,0.6}{0.1384}\\
\midrule
4DGT & $\times$ & 1 & $\times$ & 17.1689 & 0.8395 & 0.2719\\ 
\cmidrule(lr){1-7}
\multirow{2}{*}{NoPoSplat} & $\times$  & 2 & $\times$ & 22.6296 & 0.8876 & 0.1736\\ 
&  $\times$  & 2 & $\checkmark$ & 23.4321 & 0.8939 & 0.1588\\
\cmidrule(lr){1-7}
\multirow{2}{*}{AnySplat} & $\times$ & 4  & $\times$ & 23.7844 & 0.9040 & 0.1737\\ 
&  $\times$ & 4  & $\checkmark$ & 25.5875 & 0.9140 & 0.1598\\ 
\cmidrule(lr){1-7}
Ours & $\times$ & 4  & $\times$ & \cellcolor[rgb]{1.0,0.7,0.3}\bfseries{26.5138} & \cellcolor[rgb]{1,0.9,0.6}\bfseries{0.9164} & \cellcolor[rgb]{1.0,0.7,0.3}\bfseries{0.1277}\\ 
\bottomrule
\end{tabularx}
\label{table: quantitative_comparsion}
\vspace{-0.2cm}
\end{table*}

\begin{figure}[h!]
	\begin{center}
 	\setlength{\abovecaptionskip}{0.2cm}
		\includegraphics[clip, width=0.95\linewidth]{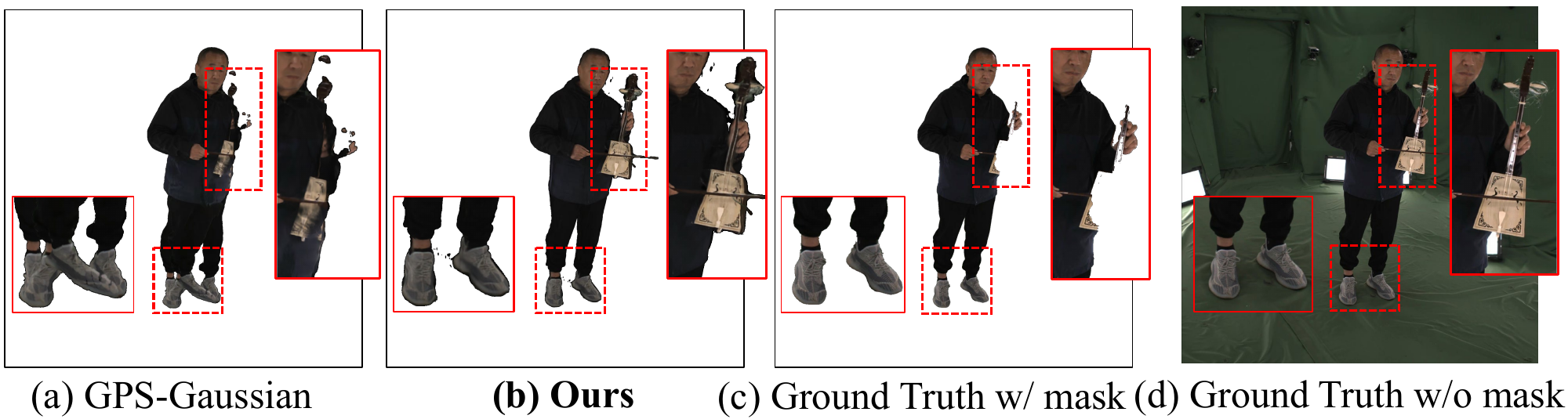}
        \caption{Pre-trained prediction masks~\cite{RVM} occasionally exhibit imperfections, introducing challenges in metric evaluation.
        (\S\ref{sec:Evaluation}). Please \faSearch ~\textbf{zoom in} to see details.}
		\label{fig:defect_on_mask}
	\end{center}
\vspace{-1.0cm}
\end{figure}

\textbf{Quantitative Evaluation.}
As demonstrated in the Table.~\ref{table: quantitative_comparsion}, our method outperforms all uncalibrated camera input approaches across all metrics. For NoPoSplat~\cite{NopoSplat} and AnySplat~\cite{AnySplat}, we optimized their camera parameters for 200 steps to better evaluate reconstruction quality, whereas our method achieves alignment with ground-truth cameras without any optimization. For AnySplat, the most directly comparable method, our approach improves PSNR by 2.7294, SSIM by 0.0124, and reduces LPIPS by 0.0460 compared to its unoptimized version, while still maintaining clear advantages over its optimized variant.
For GPS-Gaussian~\cite{GPS-Gaussian}, despite achieving comparable SSIM when provided with ground-truth cameras and eight input views, it cannot reconstruct from as few views as our method. Moreover, we observe that GPS-Gaussian's metric advantage stems from its use of masked input images, making it favorable when metrics are computed against masked ground truth. As illustrated in Fig.~\ref{fig:defect_on_mask}, our method inputs full images with background (d) and thus reconstructs the instrument region, while GPS-Gaussian's masked input (c) omits it. However, since all results are masked before evaluation against (c), GPS-Gaussian's artifact (extra feet in (a)) is suppressed, and our complete instrument reconstruction does not confer metric benefit.

\begin{table}[!ht]
    \vspace{-0.6cm}
    \centering
    \caption{\textbf{Overall computational consumption}(\S\ref{sec:Ablation Study}).} 
    \label{table: computational_consumption}
    \vspace{-0.4cm} 
    \begin{subtable}[t][0.2\textheight][t]{0.48\linewidth}
        \centering
        \caption{Training consumption}
        \label{table: training}
        \renewcommand{\arraystretch}{1.0}
        \begin{tabularx}{\linewidth}{@{\hspace{0.1cm}}>{\centering\arraybackslash}p{1.0cm} >{\centering\arraybackslash}p{1.2cm} >{\centering\arraybackslash}p{1.0cm} >{\centering\arraybackslash}p{1.6cm}}
            \toprule
            In. Res. & Sup. Res. & Side-tuning & VRAM (MiB) \\
            \midrule
            518 & 518 & $\checkmark$ & 40503\\ 
            518 & 2072 & $\checkmark$ & 44125\\
            2072 & 2072 & $\checkmark$ & 59095\\ 
            \midrule
            2072 & 2072 & $\times$ & OOM\\ 
            \bottomrule
        \end{tabularx}
    \end{subtable}
    \hfill
    \begin{subtable}[t][0.2\textheight][t]{0.48\linewidth}
        \centering
        \caption{Inference consumption}
        \label{table: inference}
        \renewcommand{\arraystretch}{1.0}
        \begin{tabularx}{\linewidth}{@{\hspace{0.1cm}}>{\centering\arraybackslash}p{1.0cm} >{\centering\arraybackslash}p{1.2cm} >{\centering\arraybackslash}p{1.2cm} >{\centering\arraybackslash}p{2.0cm}}
            \toprule
            In. Res. & Ren. Res. & VRAM (MiB) & Frame Rate (RTX 4090)\\
            \midrule
            518 & 2072 & 10852 & 4.40 FPS\\ 
            1036 & 2072 & 10886 & 4.02 FPS\\ 
            2072 & 2072 & 14052 & 3.01 FPS\\ 
            \bottomrule
        \end{tabularx}
    \end{subtable}
    \vspace{-1.2cm}
\end{table}

\subsection{Ablation Study}
\label{sec:Ablation Study}

\textbf{Ablation on Computational Consumption.}
As shown in the Table.~\ref{table: training}, we report input resolution, supervision resolution, and VRAM consumption across training stages. In the third row, with 2K input and 2K supervision, High-resolution Side-tuning increases VRAM usage by only 33.9\% compared to 0.5K input and supervision. Without side-tuning, directly increasing the AA Transformer input resolution exceeds 80GB VRAM, resulting in out-of-memory errors. This validates the computational efficiency of our side-tuning approach.
During inference, as shown in the Table.~\ref{table: inference}, we demonstrate the inference computational consumption using a single RTX 4090 GPU. Our method achieves 3.01 FPS when rendering at 2K resolution, only 24\% lower than the 4.40 FPS achieved at 0.5K. This performance reduction primarily stems from the increased resolution of the Gaussian renderer. Notably, when both configurations render at 2K, using 1K image input reduces speed by merely 8.6\% compared to 0.5K input.

\begin{figure}[htb] 
    \vspace{-0.4cm}
    \centering 
    \begin{minipage}{0.48\linewidth} 
        \centering 
        \setlength{\abovecaptionskip}{0.2cm}
        \includegraphics[clip, width=\linewidth]{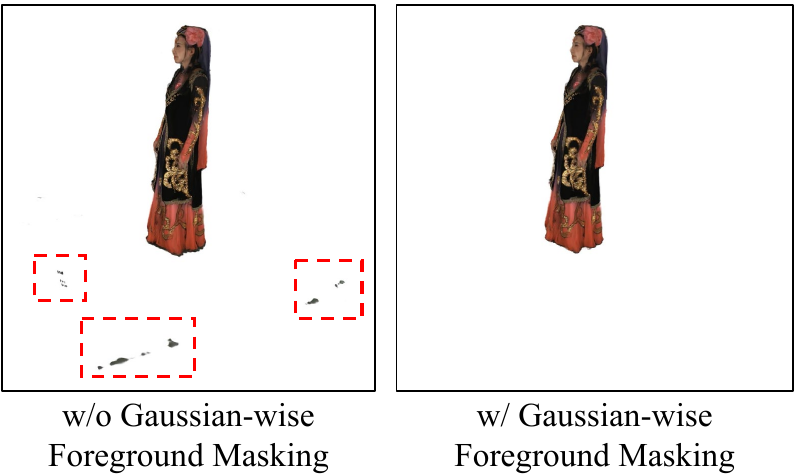} 
        \caption{\textbf{Ablation on Gaussian-wise Foreground Masking.} Our method effectively removes the background.}
        \label{fig:gs_masking}
    \end{minipage}
    \hfill 
    \begin{minipage}{0.46\linewidth} 
        \centering 
        \setlength{\abovecaptionskip}{0.2cm}
        \includegraphics[clip, width=\linewidth]{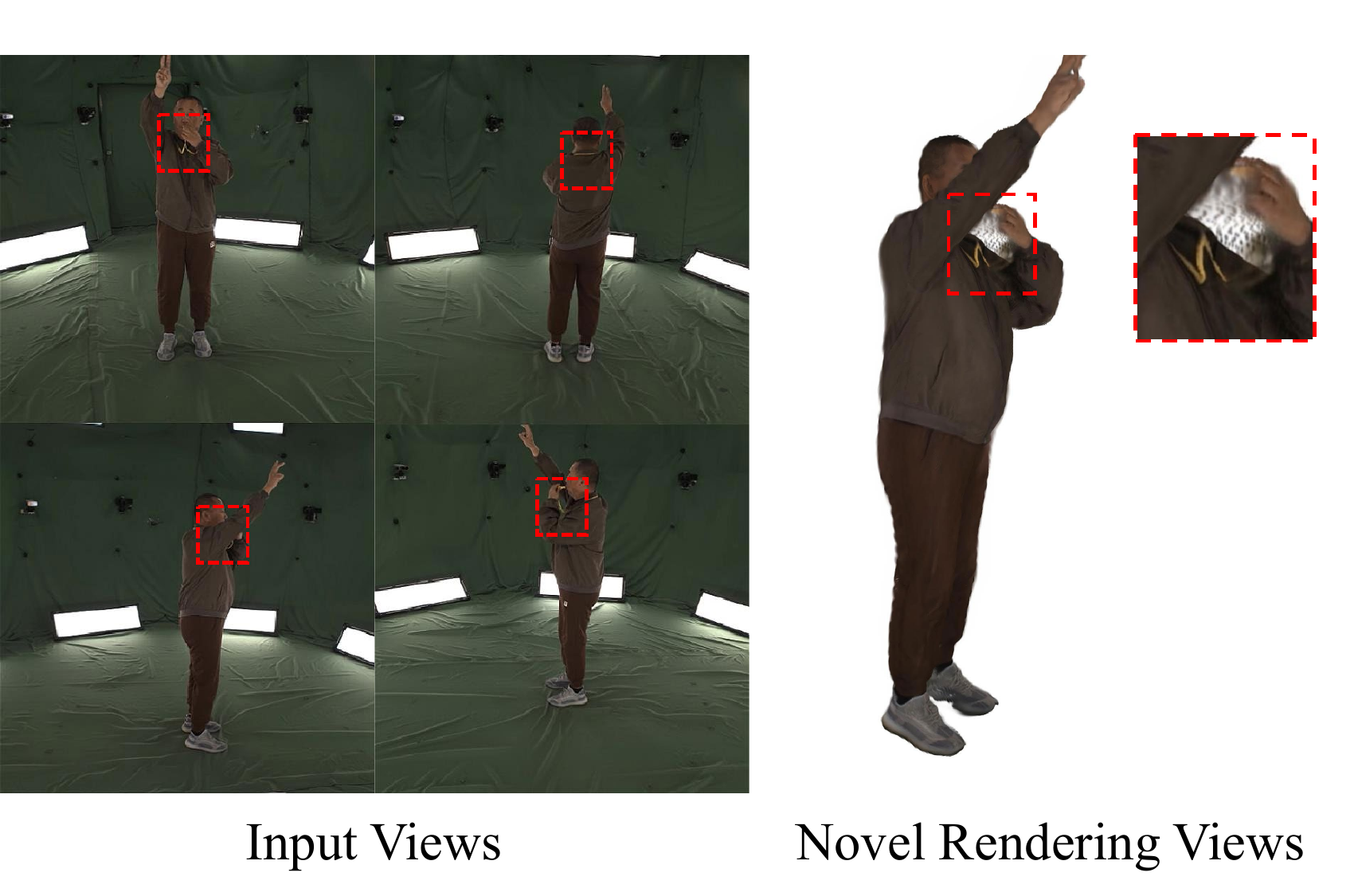} 
        \caption{\textbf{Limitation.} Our method exhibits isolated points in regions occluded from multiple viewpoints.}
        \label{fig:limitation}
    \end{minipage}
    \vspace{-0.4cm} 
\end{figure}

\textbf{Ablation on Gaussian-wise Foreground Masking.}
As shown in Fig.~\ref{fig:gs_masking}, our masking method effectively removes the background. The minimal redundant elements in w/o Masking images demonstrate that our mask has trained the Gaussian Decoder to avoid reconstructing regions beyond the mask boundary. With Gaussian-wise Foreground Masking, the decoder focuses exclusively on the foreground human subject rather than background regions during training, thereby enhancing reconstruction quality.

\section{Limitation and Future Work}
\label{sec:Limitation}

As shown in Fig.~\ref{fig:limitation}, our method exhibits isolated points in regions occluded from multiple viewpoints. Among the four input views, only the frontal view captures the region circled in red; consequently, insufficient side-view information leads to a few discrete points appearing in this area.
In the future, we plan to incorporate human priors for hands and faces to enhance geometric and topological accuracy in these regions.

\section{Conclusion}
\label{sec:Conclusion}
We present HiReFF, a feed-forward framework achieving 2K-resolution 360° human video reconstruction from uncalibrated sparse-view inputs. By decomposing the problem into temporally consistent 3D Gaussian reconstruction and efficient high-resolution synthesis, we bridge the gap between feed-forward efficiency and photorealistic volumetric video.
Our Scale-synchronized Camera Calibration and Gaussian-wise Foreground Masking resolve scale ambiguity and foreground reconstruction, while the  High-resolution Side-tuning enables 2K rendering with only modest computational overhead. Extensive experiments demonstrate state-of-the-art performance, unlocking practical applications in holographic communication, AR/VR, and live sports broadcasting without requiring calibrated rigs or per-scene optimization.

%
%

\section*{Acknowledgements}
Supported by New Generation Artificial Intelligence-National Science and Technology Major Project (2025ZD0124000). This paper is also supported by National Natural Science Foundation of China (62125107, 62572062, 62525204).

\bibliographystyle{splncs04}
\bibliography{main}

@String(CVPR= {IEEE Conf. Comput. Vis. Pattern Recog.})

@String(ICCV= {Int. Conf. Comput. Vis.})

@String(ECCV= {Eur. Conf. Comput. Vis.})

@String(TOG= {ACM Trans. Graph.})

@String(ICLR = {Int. Conf. Learn. Represent.})

@String(AAAI = {AAAI})

@String(CVPR  = {CVPR})

@String(ICCV  = {ICCV})

@String(ECCV  = {ECCV})

@String(TOG   = {ACM TOG})

@String(ICLR  = {ICLR})

@inproceedings{AnimatableGaussians,
  author       = {Zhe Li and
                  Zerong Zheng and
                  Lizhen Wang and
                  Yebin Liu},
  title        = {Animatable Gaussians: Learning Pose-Dependent Gaussian Maps for High-Fidelity
                  Human Avatar Modeling},
  booktitle    = {{IEEE/CVF} Conference on Computer Vision and Pattern Recognition},
  pages        = {19711--19722},
  year         = {2024},
}

@article{Forg4D,
  author       = {Yingdong Hu and
                  Yisheng He and
                  Jinnan Chen and
                  Weihao Yuan and
                  Kejie Qiu and
                  Zehong Lin and
                  Siyu Zhu and
                  Zilong Dong and
                  Jun Zhang},
  title        = {Forge4D: Feed-Forward 4D Human Reconstruction and Interpolation from
                  Uncalibrated Sparse-view Videos},
  journal      = {CoRR},
  volume       = {abs/2509.24209},
  year         = {2025},
  eprinttype    = {arXiv},
  eprint       = {2509.24209},
}

@inproceedings{MapAnything,
  title={MapAnything: Universal Feed-Forward Metric 3D Reconstruction; map-anything. github. io},
  author={Keetha, Nikhil and M{\"u}ller, Norman and Sch{\"o}nberger, Johannes and Porzi, Lorenzo and Zhang, Yuchen and Fischer, Tobias and Knapitsch, Arno and Zauss, Duncan and Weber, Ethan and Antunes, Nelson and others},
  booktitle={2026 International Conference on 3D Vision (3DV)},
  pages={499--509},
  year={2026},
  organization={IEEE}
}

@article{GaussianSR,
  author       = {Xiqian Yu and
                  Hanxin Zhu and
                  Tianyu He and
                  Zhibo Chen},
  title        = {GaussianSR: 3D Gaussian Super-Resolution with 2D Diffusion Priors},
  journal      = {CoRR},
  volume       = {abs/2406.10111},
  year         = {2024},
  eprinttype    = {arXiv},
  eprint       = {2406.10111}
}

@inproceedings{FastAvatar,
  title={Fastavatar: Towards unified and fast 3d avatar reconstruction with large gaussian reconstruction transformers},
  author={Wu, Yue and Chen, Xuanhong and Wu, Yufan and Li, Wen and Lu, Yuxi and Feng, Kairui},
  booktitle={The Fourteenth International Conference on Learning Representations},
  year={2026}
}

@article{AnySplat,
  title={Anysplat: Feed-forward 3d gaussian splatting from unconstrained views},
  author={Jiang, Lihan and Mao, Yucheng and Xu, Linning and Lu, Tao and Ren, Kerui and Jin, Yichen and Xu, Xudong and Yu, Mulin and Pang, Jiangmiao and Zhao, Feng and others},
  journal={ACM Transactions on Graphics (TOG)},
  volume={44},
  number={6},
  pages={1--16},
  year={2025}
}

@inproceedings{VGGT,
  author       = {Jianyuan Wang and
                  Minghao Chen and
                  Nikita Karaev and
                  Andrea Vedaldi and
                  Christian Rupprecht and
                  David Novotn{\'{y}}},
  title        = {{VGGT:} Visual Geometry Grounded Transformer},
  booktitle    = {{IEEE/CVF} Conference on Computer Vision and Pattern Recognition},
  pages        = {5294--5306},
  year         = {2025},
}

@article{Pi3,
  author       = {Yifan Wang and
                  Jianjun Zhou and
                  Haoyi Zhu and
                  Wenzheng Chang and
                  Yang Zhou and
                  Zizun Li and
                  Junyi Chen and
                  Jiangmiao Pang and
                  Chunhua Shen and
                  Tong He},
  title        = {{\(\pi\)}\({}^{\mbox{3}}\): Scalable Permutation-Equivariant Visual
                  Geometry Learning},
  journal      = {CoRR},
  volume       = {abs/2507.13347},
  year         = {2025},
  eprinttype    = {arXiv},
  eprint       = {2507.13347},
}

@inproceedings{NopoSplat,
  author       = {Botao Ye and
                  Sifei Liu and
                  Haofei Xu and
                  Xueting Li and
                  Marc Pollefeys and
                  Ming{-}Hsuan Yang and
                  Songyou Peng},
  title        = {No Pose, No Problem: Surprisingly Simple 3D Gaussian Splats from Sparse
                  Unposed Images},
  booktitle    = {The Thirteenth International Conference on Learning Representations,
                  {ICLR}},
  year         = {2025}
}

@inproceedings{GPS-Gaussian,
  author       = {Shunyuan Zheng and
                  Boyao Zhou and
                  Ruizhi Shao and
                  Boning Liu and
                  Shengping Zhang and
                  Liqiang Nie and
                  Yebin Liu},
  title        = {GPS-Gaussian: Generalizable Pixel-Wise 3D Gaussian Splatting for Real-Time
                  Human Novel View Synthesis},
  booktitle    = {{IEEE/CVF} Conference on Computer Vision and Pattern Recognition},
  pages        = {19680--19690},
  year         = {2024},
}

@inproceedings{DUSt3R,
  author       = {Shuzhe Wang and
                  Vincent Leroy and
                  Yohann Cabon and
                  Boris Chidlovskii and
                  J{\'{e}}r{\^{o}}me Revaud},
  title        = {DUSt3R: Geometric 3D Vision Made Easy},
  booktitle    = {{IEEE/CVF} Conference on Computer Vision and Pattern Recognition},
  pages        = {20697--20709},
  year         = {2024},
}

@inproceedings{4DGT,
  title={4DGT: Learning a 4D Gaussian Transformer Using Real-World Monocular Videos},
  author={Xu, Zhen and Li, Zhengqin and Dong, Zhao and Zhou, Xiaowei and Newcombe, Richard and Lv, Zhaoyang},
  booktitle={The Thirty-ninth Annual Conference on Neural Information Processing Systems}
}

@inproceedings{Diffuman4D,
  title={Diffuman4d: 4d consistent human view synthesis from sparse-view videos with spatio-temporal diffusion models},
  author={Jin, Yudong and Peng, Sida and Wang, Xuan and Xie, Tao and Xu, Zhen and Yang, Yifan and Shen, Yujun and Bao, Hujun and Zhou, Xiaowei},
  booktitle={Proceedings of the IEEE/CVF International Conference on Computer Vision},
  pages={11047--11057},
  year={2025}
}

@inproceedings{DoubleField,
  author       = {Ruizhi Shao and
                  Hongwen Zhang and
                  He Zhang and
                  Mingjia Chen and
                  Yanpei Cao and
                  Tao Yu and
                  Yebin Liu},
  title        = {DoubleField: Bridging the Neural Surface and Radiance Fields for High-fidelity
                  Human Reconstruction and Rendering},
  booktitle    = {{IEEE/CVF} Conference on Computer Vision and Pattern Recognition},
  pages        = {15851--15861},
  year         = {2022},
}

@ARTICLE{HFHuman,
  author={Jiang, Yiming and Song, Wenfeng and Li, Shuai and Hao, Aimin},
  journal={IEEE Transactions on Visualization and Computer Graphics}, 
  title={HFHuman: High-Fidelity Human Reconstruction From Single Image With Multi-Modality Fusion}, 
  year={2026},
  volume={32},
  number={2},
  pages={2152-2164},
}

@inproceedings{Gbc-splat,
  author       = {Hanzhang Tu and
                  Zhanfeng Liao and
                  Boyao Zhou and
                  Shunyuan Zheng and
                  Xilong Zhou and
                  Liuxin Zhang and
                  QianYing Wang and
                  Yebin Liu},
  title        = {GBC-Splat: Generalizable Gaussian-Based Clothed Human Digitalization
                  under Sparse {RGB} Cameras},
  booktitle    = {{IEEE/CVF} Conference on Computer Vision and Pattern Recognition},
  pages        = {26377--26387},
  year         = {2025}
}

@inproceedings{MVSplat,
  author       = {Yuedong Chen and
                  Haofei Xu and
                  Chuanxia Zheng and
                  Bohan Zhuang and
                  Marc Pollefeys and
                  Andreas Geiger and
                  Tat{-}Jen Cham and
                  Jianfei Cai},
  editor       = {Ales Leonardis and
                  Elisa Ricci and
                  Stefan Roth and
                  Olga Russakovsky and
                  Torsten Sattler and
                  G{\"{u}}l Varol},
  title        = {MVSplat: Efficient 3D Gaussian Splatting from Sparse Multi-view Images},
  booktitle    = {Computer Vision - {ECCV} 2024 - 18th European Conference},
  volume       = {15079},
  pages        = {370--386},
  year         = {2024},
}

@inproceedings{HGM,
  author       = {Jinnan Chen and
                  Chen Li and
                  Jianfeng Zhang and
                  Lingting Zhu and
                  Buzhen Huang and
                  Hanlin Chen and
                  Gim Hee Lee},
  title        = {Generalizable Human Gaussians from Single-View Image},
  booktitle    = {The Thirteenth International Conference on Learning Representations,
                  {ICLR}},
  year         = {2025}
}

@inproceedings{FLARE,
  author       = {Shangzhan Zhang and
                  Jianyuan Wang and
                  Yinghao Xu and
                  Nan Xue and
                  Christian Rupprecht and
                  Xiaowei Zhou and
                  Yujun Shen and
                  Gordon Wetzstein},
  title        = {{FLARE:} Feed-forward Geometry, Appearance and Camera Estimation from
                  Uncalibrated Sparse Views},
  booktitle    = {{IEEE/CVF} Conference on Computer Vision and Pattern Recognition},
  pages        = {21936--21947},
  year         = {2025},
}

@inproceedings{Tele-Aloha,
  author       = {Hanzhang Tu and
                  Ruizhi Shao and
                  Xue Dong and
                  Shunyuan Zheng and
                  Hao Zhang and
                  Lili Chen and
                  Meili Wang and
                  Wenyu Li and
                  Siyan Ma and
                  Shengping Zhang and
                  Boyao Zhou and
                  Yebin Liu},
  editor       = {Andres Burbano and
                  Denis Zorin and
                  Wojciech Jarosz},
  title        = {Tele-Aloha: {A} Telepresence System with Low-budget and High-authenticity
                  Using Sparse {RGB} Cameras},
  booktitle    = {{ACM} {SIGGRAPH} 2024 Conference Papers},
  pages        = {116},
  year         = {2024}
}

@article{SRGS,
  author       = {Xiang Feng and
                  Yongbo He and
                  Yubo Wang and
                  Yan Yang and
                  Zhenzhong Kuang and
                  Jun Yu and
                  Jianping Fan and
                  Jiajun Ding},
  title        = {{SRGS:} Super-Resolution 3D Gaussian Splatting},
  journal      = {CoRR},
  volume       = {abs/2404.10318},
  year         = {2024},
  eprinttype    = {arXiv},
  eprint       = {2404.10318}
}

@inproceedings{Arbitrary-ScaleGS,
  title={Arbitrary-Scale 3D Gaussian Super-Resolution},
  author={Zeng, Huimin and Bai, Yue and Fu, Yun},
  booktitle={Proceedings of the AAAI Conference on Artificial Intelligence},
  volume={40},
  number={15},
  pages={12304--12312},
  year={2026}
}

@inproceedings{Mip-Splatting,
  author       = {Zehao Yu and
                  Anpei Chen and
                  Binbin Huang and
                  Torsten Sattler and
                  Andreas Geiger},
  title        = {Mip-Splatting: Alias-Free 3D Gaussian Splatting},
  booktitle    = {{IEEE/CVF} Conference on Computer Vision and Pattern Recognition},
  pages        = {19447--19456},
  year         = {2024},
}

@inproceedings{NeRF-SR,
  author       = {Chen Wang and
                  Xian Wu and
                  Yuan{-}Chen Guo and
                  Song{-}Hai Zhang and
                  Yu{-}Wing Tai and
                  Shi{-}Min Hu},
  editor       = {Jo{\~{a}}o Magalh{\~{a}}es and
                  Alberto Del Bimbo and
                  Shin'ichi Satoh and
                  Nicu Sebe and
                  Xavier Alameda{-}Pineda and
                  Qin Jin and
                  Vincent Oria and
                  Laura Toni},
  title        = {NeRF-SR: High Quality Neural Radiance Fields using Supersampling},
  booktitle    = {{MM} '22: The 30th {ACM} International Conference on Multimedia, Lisboa,
                  Portugal, October 10 - 14, 2022},
  pages        = {6445--6454},
  year         = {2022},
}

@article{Super-NeRF,
  author       = {Yuqi Han and
                  Tao Yu and
                  Xiaohang Yu and
                  Di Xu and
                  Binge Zheng and
                  Zonghong Dai and
                  Changpeng Yang and
                  Yuwang Wang and
                  Qionghai Dai},
  title        = {Super-NeRF: View-Consistent Detail Generation for NeRF Super-Resolution},
  journal      = {{IEEE} Trans. Vis. Comput. Graph.},
  volume       = {31},
  number       = {9},
  pages        = {6053--6066},
  year         = {2025},
}

@inproceedings{RefSR-NeRF,
  author       = {Xudong Huang and
                  Wei Li and
                  Jie Hu and
                  Hanting Chen and
                  Yunhe Wang},
  title        = {RefSR-NeRF: Towards High Fidelity and Super Resolution View Synthesis},
  booktitle    = {{IEEE/CVF} Conference on Computer Vision and Pattern Recognition},
  pages        = {8244--8253},
  year         = {2023},
}

@inproceedings{Side-Tuning,
  author       = {Jeffrey O. Zhang and
                  Alexander Sax and
                  Amir Zamir and
                  Leonidas J. Guibas and
                  Jitendra Malik},
  editor       = {Andrea Vedaldi and
                  Horst Bischof and
                  Thomas Brox and
                  Jan{-}Michael Frahm},
  title        = {Side-Tuning: {A} Baseline for Network Adaptation via Additive Side
                  Networks},
  booktitle    = {Computer Vision - {ECCV} 2020 - 16th European Conference},
  volume       = {12348},
  pages        = {698--714},
  year         = {2020},
}

@article{DPT,
  author       = {Ren{\'{e}} Ranftl and
                  Katrin Lasinger and
                  David Hafner and
                  Konrad Schindler and
                  Vladlen Koltun},
  title        = {Towards Robust Monocular Depth Estimation: Mixing Datasets for Zero-Shot
                  Cross-Dataset Transfer},
  journal      = {{IEEE} Trans. Pattern Anal. Mach. Intell.},
  volume       = {44},
  number       = {3},
  pages        = {1623--1637},
  year         = {2022},
}

@inproceedings{DNA-Rendering,
  author       = {Wei Cheng and
                  Ruixiang Chen and
                  Siming Fan and
                  Wanqi Yin and
                  Keyu Chen and
                  Zhongang Cai and
                  Jingbo Wang and
                  Yang Gao and
                  Zhengming Yu and
                  Zhengyu Lin and
                  Daxuan Ren and
                  Lei Yang and
                  Ziwei Liu and
                  Chen Change Loy and
                  Chen Qian and
                  Wayne Wu and
                  Dahua Lin and
                  Bo Dai and
                  Kwan{-}Yee Lin},
  title        = {DNA-Rendering: {A} Diverse Neural Actor Repository for High-Fidelity
                  Human-centric Rendering},
  booktitle    = {{IEEE/CVF} International Conference on Computer Vision, {ICCV}},
  pages        = {19925--19936},
  year         = {2023},
}

@inproceedings{LHM,
  title={LHM: Large Animatable Human Reconstruction Model for Single Image to 3D in Seconds},
  author={Qiu, Lingteng and Gu, Xiaodong and Li, Peihao and Zuo, Qi and Shen, Weichao and Zhang, Junfei and Qiu, Kejie and Yuan, Weihao and Chen, Guanying and Dong, Zilong and others},
  booktitle={Proceedings of the IEEE/CVF International Conference on Computer Vision},
  pages={14184--14194},
  year={2025}
}

@inproceedings{PSHuman,
  author       = {Peng Li and
                  Wangguandong Zheng and
                  Yuan Liu and
                  Tao Yu and
                  Yangguang Li and
                  Xingqun Qi and
                  Xiaowei Chi and
                  Siyu Xia and
                  Yan{-}Pei Cao and
                  Wei Xue and
                  Wenhan Luo and
                  Yike Guo},
  title        = {PSHuman: Photorealistic Single-image 3D Human Reconstruction using
                  Cross-Scale Multiview Diffusion and Explicit Remeshing},
  booktitle    = {{IEEE/CVF} Conference on Computer Vision and Pattern Recognition},
  pages        = {16008--16018},
  year         = {2025},
}

@inproceedings{Magicman,
  title={Magicman: Generative novel view synthesis of humans with 3d-aware diffusion and iterative refinement},
  author={He, Xu and Wu, Zhiyong and Li, Xiaoyu and Kang, Di and Zhang, Chaopeng and Ye, Jiangnan and Chen, Liyang and Gao, Xiangjun and Zhang, Han and Zhuang, Haolin},
  booktitle={Proceedings of the AAAI Conference on Artificial Intelligence},
  volume={39},
  number={3},
  pages={3437--3445},
  year={2025}
}

@inproceedings{Splat-SAP,
  title={Splat-SAP: Feed-Forward Gaussian Splatting for Human-Centered Scene with
Scale-Aware Point Map Reconstruction},
  author={Zhou, Boyao and Zheng, Shunyuan and Liao, Zhanfeng and Ma, Zihan and Tu, Hanzhang and Liu, Boning and Liu, Yebin},
  booktitle={Proceedings of the AAAI Conference on Artificial Intelligence},
  year={2026}
}

@article{DECON,
   title={DECON: Reconstruction of Clothed-Geometric Multiple Humans from a Single Image via Geometry-Guided Decoupling},
   volume={40},
   number={7},
   journal={Proceedings of the AAAI Conference on Artificial Intelligence},
   author={Jiang, Yiming and Song, Wenfeng and Li, Shuai and Hao, Aimin},
   year={2026},
   pages={5459-5467}
}

@inproceedings{Idol,
  title={Idol: Instant photorealistic 3d human creation from a single image},
  author={Zhuang, Yiyu and Lv, Jiaxi and Wen, Hao and Shuai, Qing and Zeng, Ailing and Zhu, Hao and Chen, Shifeng and Yang, Yujiu and Cao, Xun and Liu, Wei},
  booktitle={Proceedings of the Computer Vision and Pattern Recognition Conference},
  pages={26308--26319},
  year={2025}
}

@inproceedings{SIFU,
  author       = {Zechuan Zhang and
                  Zongxin Yang and
                  Yi Yang},
  title        = {{SIFU:} Side-view Conditioned Implicit Function for Real-world Usable
                  Clothed Human Reconstruction},
  booktitle    = {{IEEE/CVF} Conference on Computer Vision and Pattern Recognition},
  pages        = {9936--9947},
  year         = {2024},
}

@article{3D_Facial,
  author       = {Wenfeng Song and
                  Xuan Wang and
                  Yiming Jiang and
                  Shuai Li and
                  Aimin Hao and
                  Xia Hou and
                  Hong Qin},
  title        = {Expressive 3D Facial Animation Generation Based on Local-to-Global
                  Latent Diffusion},
  journal      = {{IEEE} Trans. Vis. Comput. Graph.},
  volume       = {30},
  number       = {11},
  pages        = {7397--7407},
  year         = {2024},
}

@inproceedings{Human_Sports,
  author       = {Tobias Baumgartner and
                  Stefanie Klatt},
  title        = {Monocular 3D Human Pose Estimation for Sports Broadcasts using Partial
                  Sports Field Registration},
  booktitle    = {{IEEE/CVF} Conference on Computer Vision and Pattern Recognition},
  pages        = {5109--5118},
  year         = {2023},
}

@inproceedings{EfficientHuman,
  title={EfficientHuman: Efficient Training and Reconstruction of Moving Human using Articulated 2D Gaussian},
  author={Tian, Hao and Liu, Rui and Shen, Wen and Hu, Yilong and Zheng, Zhihao and Qin, Xiaolin},
  booktitle={2025 International Joint Conference on Neural Networks (IJCNN)},
  pages={1--8},
  year={2025},
  organization={IEEE}
}

@article{Human3R,
    title={Human3R: Everyone Everywhere All at Once},
    author={Chen, Yue and Chen, Xingyu and Xue, Yuxuan and Chen, Anpei and Xiu, Yuliang and Gerard, Pons-Moll},
    journal={arXiv preprint arXiv:2510.06219},
    year={2025}
    }

@article{StreamVGGT,
  title={Streaming 4d visual geometry transformer},
  author={Zhuo, Dong and Zheng, Wenzhao and Guo, Jiahe and Wu, Yuqi and Zhou, Jie and Lu, Jiwen},
  journal={arXiv preprint arXiv:2507.11539},
  year={2025}
}

@inproceedings{CAT4D,
  author       = {Rundi Wu and
                  Ruiqi Gao and
                  Ben Poole and
                  Alex Trevithick and
                  Changxi Zheng and
                  Jonathan T. Barron and
                  Aleksander Holynski},
  title        = {{CAT4D:} Create Anything in 4D with Multi-View Video Diffusion Models},
  booktitle    = {{IEEE/CVF} Conference on Computer Vision and Pattern Recognition},
  pages        = {26057--26068},
  year         = {2025},
}

@article{Longvolcap,
  author       = {Zhen Xu and
                  Yinghao Xu and
                  Zhiyuan Yu and
                  Sida Peng and
                  Jiaming Sun and
                  Hujun Bao and
                  Xiaowei Zhou},
  title        = {Representing Long Volumetric Video with Temporal Gaussian Hierarchy},
  journal      = {{ACM} Trans. Graph.},
  volume       = {43},
  number       = {6},
  pages        = {171:1--171:18},
  year         = {2024},
}

@article{Lyra,
    title={Lyra: Generative 3D Scene Reconstruction via Self-Distillation with Video Diffusion Models},
    author={Bahmani, Sherwin and Shen, Tianchang and Ren, Jiawei and Huang, Jiahui and Jiang, Yifeng and Turki, Haithem and Tagliasacchi, Andrea and Lindell, David B. and Gojcic, Zan and Fidler, Sanja and Ling, Huan and Gao, Jun and Ren, Xuanchi},
    journal={arXiv preprint arXiv:2509.19296},
    year={2025}
}

@article{4DNeX,
    title={4DNeX: Feed-Forward 4D Generative Modeling Made Easy},
    author={Chen, Zhaoxi and Liu, Tianqi and Zhuo, Long and Ren, Jiawei and Tao, Zeng and Zhu, He and Hong, Fangzhou and Pan, Liang and Liu, Ziwei},
    journal={arXiv preprint arXiv:2508.13154},
    year={2025}
  }

@misc{FlashWorld,
        title={FlashWorld: High-quality 3D Scene Generation within Seconds},
        author={Xinyang Li and Tengfei Wang and Zixiao Gu and Shengchuan Zhang and Chunchao Guo and Liujuan Cao},
        journal={arXiv preprint arXiv:2510.13678},
        year={2025},
        eprint={2510.13678},
        archivePrefix={arXiv},
        primaryClass={cs.CV}
    }

@article{VolSplat,
  title={VolSplat: Rethinking Feed-Forward 3D Gaussian Splatting with Voxel-Aligned Prediction},
  author={Wang, Weijie and Chen, Yeqing and Zhang, Zeyu and Liu, Hengyu and Wang, Haoxiao and Feng, Zhiyuan and Qin, Wenkang and Zhu, Zheng and Chen, Donny Y. and Zhuang, Bohan},
  journal={arXiv preprint arXiv:2509.19297},
  year={2025}
}

@inproceedings{DepthSplat,
  author       = {Haofei Xu and
                  Songyou Peng and
                  Fangjinhua Wang and
                  Hermann Blum and
                  Daniel Barath and
                  Andreas Geiger and
                  Marc Pollefeys},
  title        = {DepthSplat: Connecting Gaussian Splatting and Depth},
  booktitle    = {{IEEE/CVF} Conference on Computer Vision and Pattern Recognition},
  pages        = {16453--16463},
  year         = {2025},
}

@inproceedings{FreeSplat,
  author       = {Yunsong Wang and
                  Tianxin Huang and
                  Hanlin Chen and
                  Gim Hee Lee},
  editor       = {Amir Globersons and
                  Lester Mackey and
                  Danielle Belgrave and
                  Angela Fan and
                  Ulrich Paquet and
                  Jakub M. Tomczak and
                  Cheng Zhang},
  title        = {FreeSplat: Generalizable 3D Gaussian Splatting Towards Free View Synthesis
                  of Indoor Scenes},
  booktitle    = {Advances in Neural Information Processing Systems 38: Annual Conference
                  on Neural Information Processing Systems},
  year         = {2024},
}

@inproceedings{HumanSplat,
  author       = {Panwang Pan and
                  Zhuo Su and
                  Chenguo Lin and
                  Zhen Fan and
                  Yongjie Zhang and
                  Zeming Li and
                  Tingting Shen and
                  Yadong Mu and
                  Yebin Liu},
  editor       = {Amir Globersons and
                  Lester Mackey and
                  Danielle Belgrave and
                  Angela Fan and
                  Ulrich Paquet and
                  Jakub M. Tomczak and
                  Cheng Zhang},
  title        = {HumanSplat: Generalizable Single-Image Human Gaussian Splatting with
                  Structure Priors},
  booktitle    = {Advances in Neural Information Processing Systems 38: Annual Conference
                  on Neural Information Processing Systems},
  year         = {2024},
}

@article{GGN,
  title={Gaussian graph network: Learning efficient and generalizable gaussian representations from multi-view images},
  author={Zhang, Shengjun and Fei, Xin and Liu, Fangfu and Song, Haixu and Duan, Yueqi},
  journal={Advances in Neural Information Processing Systems},
  volume={37},
  pages={50361--50380},
  year={2024}
}

@misc{FastVGGT,
      title={FastVGGT: Training-Free Acceleration of Visual Geometry Transformer}, 
      author={You Shen and Zhipeng Zhang and Yansong Qu and Liujuan Cao},
      journal={arXiv preprint arXiv:2509.02560},
      year={2025},
      eprint={2509.02560},
      archivePrefix={arXiv},
      primaryClass={cs.CV}
}

@inproceedings{Avat3r,
  title={Avat3r: Large animatable gaussian reconstruction model for high-fidelity 3d head avatars},
  author={Kirschstein, Tobias and Romero, Javier and Sevastopolsky, Artem and Nie{\ss}ner, Matthias and Saito, Shunsuke},
  booktitle={Proceedings of the IEEE/CVF International Conference on Computer Vision},
  pages={12089--12100},
  year={2025}
}

@article{gsplat,
  title={gsplat: An open-source library for Gaussian splatting},
  author={Ye, Vickie and Li, Ruilong and Kerr, Justin and Turkulainen, Matias and Yi, Brent and Pan, Zhuoyang and Seiskari, Otto and Ye, Jianbo and Hu, Jeffrey and Tancik, Matthew and Angjoo Kanazawa},
  journal={Journal of Machine Learning Research},
  volume={26},
  number={34},
  pages={1--17},
  year={2025}
}

@inproceedings{Perceptual_Loss,
  author       = {Justin Johnson and
                  Alexandre Alahi and
                  Li Fei{-}Fei},
  editor       = {Bastian Leibe and
                  Jiri Matas and
                  Nicu Sebe and
                  Max Welling},
  title        = {Perceptual Losses for Real-Time Style Transfer and Super-Resolution},
  booktitle    = {Computer Vision - {ECCV} 2016 - 14th European Conference},
  volume       = {9906},
  pages        = {694--711},
  year         = {2016},
}

@inproceedings{VGG,
  author       = {Karen Simonyan and
                  Andrew Zisserman},
  editor       = {Yoshua Bengio and
                  Yann LeCun},
  title        = {Very Deep Convolutional Networks for Large-Scale Image Recognition},
  booktitle    = {3rd International Conference on Learning Representations, Conference Track Proceedings},
  year         = {2015},
}

@inproceedings{DiffuStereo,
  author       = {Ruizhi Shao and
                  Zerong Zheng and
                  Hongwen Zhang and
                  Jingxiang Sun and
                  Yebin Liu},
  editor       = {Shai Avidan and
                  Gabriel J. Brostow and
                  Moustapha Ciss{\'{e}} and
                  Giovanni Maria Farinella and
                  Tal Hassner},
  title        = {DiffuStereo: High Quality Human Reconstruction via Diffusion-Based
                  Stereo Using Sparse Cameras},
  booktitle    = {Computer Vision - {ECCV} 2022 - 17th European Conference},
  volume       = {13692},
  pages        = {702--720},
  year         = {2022},
}

@article{Vid2Actor,
  author       = {Chung{-}Yi Weng and
                  Brian Curless and
                  Ira Kemelmacher{-}Shlizerman},
  title        = {Vid2Actor: Free-viewpoint Animatable Person Synthesis from Video in
                  the Wild},
  journal      = {CoRR},
  volume       = {abs/2012.12884},
  year         = {2020},
  eprinttype    = {arXiv},
  eprint       = {2012.12884},
}

@inproceedings{AniGS,
  author       = {Lingteng Qiu and
                  Shenhao Zhu and
                  Qi Zuo and
                  Xiaodong Gu and
                  Yuan Dong and
                  Junfei Zhang and
                  Chao Xu and
                  Zhe Li and
                  Weihao Yuan and
                  Liefeng Bo and
                  Guanying Chen and
                  Zilong Dong},
  title        = {AniGS: Animatable Gaussian Avatar from a Single Image with Inconsistent
                  Gaussian Reconstruction},
  booktitle    = {{IEEE/CVF} Conference on Computer Vision and Pattern Recognition},
  pages        = {21148--21158},
  year         = {2025}
}

@inproceedings{CtrlAvatar,
  author       = {Wenfeng Song and
                  Yang Ding and
                  Fei Hou and
                  Shuai Li and
                  Aimin Hao and
                  Xia Hou},
  editor       = {Toby Walsh and
                  Julie Shah and
                  Zico Kolter},
  title        = {CtrlAvatar: Controllable Avatars Generation via Disentangled Invertible
                  Networks},
  booktitle    = {AAAI-25, Sponsored by the Association for the Advancement of Artificial
                  Intelligence},
  pages        = {6959--6967},
  year         = {2025}
}

@article{SMPL,
  author       = {Matthew Loper and
                  Naureen Mahmood and
                  Javier Romero and
                  Gerard Pons{-}Moll and
                  Michael J. Black},
  title        = {{SMPL:} a skinned multi-person linear model},
  journal      = {{ACM} Trans. Graph.},
  volume       = {34},
  number       = {6},
  pages        = {248:1--248:16},
  year         = {2015},
}

@InProceedings{HumanNeRF,
    author    = {Zhao, Fuqiang and Yang, Wei and Zhang, Jiakai and Lin, Pei and Zhang, Yingliang and Yu, Jingyi and Xu, Lan},
    title     = {HumanNeRF: Efficiently Generated Human Radiance Field From Sparse Inputs},
    booktitle = {Proceedings of the IEEE/CVF Conference on Computer Vision and Pattern Recognition (CVPR)},
    month     = {June},
    year      = {2022},
    pages     = {7743-7753}
}

@inproceedings{3dgs-avatar,
  title={3dgs-avatar: Animatable avatars via deformable 3d gaussian splatting},
  author={Qian, Zhiyin and Wang, Shaofei and Mihajlovic, Marko and Geiger, Andreas and Tang, Siyu},
  booktitle={Proceedings of the IEEE/CVF conference on computer vision and pattern recognition},
  pages={5020--5030},
  year={2024}
}

@inproceedings{RoGSplat,
  title={RoGSplat: Learning Robust Generalizable Human Gaussian Splatting from Sparse Multi-View Images},
  author={Xiao, Junjin and Zhang, Qing and Nie, Yonewei and Zhu, Lei and Zheng, Wei-Shi},
  booktitle={Proceedings of the Computer Vision and Pattern Recognition Conference},
  pages={5980--5990},
  year={2025}
}

@article{Human4DiT,
    title={Human4DiT: 360-degree Human Video Generation with 4D Diffusion Transformer},
    author={Shao, Ruizhi and Pang, Youxin and Zheng, Zerong and Sun, Jingxiang and Liu, Yebin},
    journal={ACM Transactions on Graphics (TOG)},
    volume={43},
    number={6},
    articleno={},
    year={2024}
}

@inproceedings{RVM,
  title={Robust high-resolution video matting with temporal guidance},
  author={Lin, Shanchuan and Yang, Linjie and Saleemi, Imran and Sengupta, Soumyadip},
  booktitle={Proceedings of the IEEE/CVF Winter Conference on Applications of Computer Vision},
  pages={238--247},
  year={2022}
}

@inproceedings{Maaz2022EdgeNeXt,
      title={EdgeNeXt: Efficiently Amalgamated CNN-Transformer Architecture for Mobile Vision Applications},
        author={Muhammad Maaz and Abdelrahman Shaker and Hisham Cholakkal and Salman Khan and Syed Waqas Zamir and Rao Muhammad Anwer and Fahad Shahbaz Khan},
      booktitle={International Workshop on Computational Aspects of Deep Learning at 17th European Conference on Computer Vision (CADL2022)},
      year={2022},
      organization={Springer}
}

@inproceedings{HumanPro,
  title={HumanPro: Single-view 3D Clothed Human Reconstruction with Progressive Normal Guidance},
  author={Sun, Jianchi and Luo, Fei and Fan, Wenzhuo and Jiang, Yu and Xiao, Chunxia},
  booktitle={Proceedings of the AAAI Conference on Artificial Intelligence},
  volume={40},
  number={11},
  pages={9180--9188},
  year={2026}
}

@article{mescheder2025sharp,
  title={Sharp monocular view synthesis in less than a second},
  author={Mescheder, Lars and Dong, Wei and Li, Shiwei and Bai, Xuyang and Santos, Marcel and Hu, Peiyun and Lecouat, Bruno and Zhen, Mingmin and Delaunoy, Ama{\~A}{\c{G}}l and Fang, Tian and others},
  journal={arXiv preprint arXiv:2512.10685},
  year={2025}
}

@inproceedings{peng2021neural,
    title={Neural Body: Implicit Neural Representations with Structured Latent Codes for Novel View Synthesis of Dynamic Humans},
    author={Peng, Sida and Zhang, Yuanqing and Xu, Yinghao and Wang, Qianqian and Shuai, Qing and Bao, Hujun and Zhou, Xiaowei},
    booktitle={CVPR},
    year={2021}
}

@inproceedings{fang2021mirrored,
    title={Reconstructing 3D Human Pose by Watching Humans in the Mirror},
    author={Fang, Qi and Shuai, Qing and Dong, Junting and Bao, Hujun and Zhou, Xiaowei},
    booktitle={CVPR},
    year={2021}
}

@inproceedings{xiong2024mvhumannet,
  title={MVHumanNet: A Large-scale Dataset of Multi-view Daily Dressing Human Captures},
  author={Xiong, Zhangyang and Li, Chenghong and Liu, Kenkun and Liao, Hongjie and Hu, Jianqiao and Zhu, Junyi and Ning, Shuliang and Qiu, Lingteng and Wang, Chongjie and Wang, Shijie and others},
  booktitle={Proceedings of the IEEE/CVF Conference on Computer Vision and Pattern Recognition},
  pages={19801--19811},
  year={2024}
}

@article{li2025mvhumannet++,
  title={MVHumanNet++: A Large-scale Dataset of Multi-view Daily Dressing Human Captures with Richer Annotations for 3D Human Digitization},
  author={Li, Chenghong and Liao, Hongjie and Zhi, Yihao and Yang, Xihe and Sun, Zhengwentai and Chang, Jiahao and Cui, Shuguang and Han, Xiaoguang},
  journal={arXiv preprint arXiv:2505.01838},
  year={2025}
}

@INPROCEEDINGS{deng2024compact,
  author={Deng, Tianchen and Chen, Yaohui and Yang, Jianfei and Yuan, Shenghai and Liu, Jiuming and Wang, Danwei and Chen, Weidong},
  booktitle={2025 IEEE/RSJ International Conference on Intelligent Robots and Systems (IROS)}, 
  title={CGS-SLAM: Compact 3D Gaussian Splatting for Dense Visual SLAM}, 
  year={2025},
  volume={},
  number={},
  pages={1606-1613}}

@InProceedings{Deng_2026_CVPR,
    author    = {Deng, Tianchen and Chen, Xuefeng and Chen, Yi and Chen, Qu and Xu, Yuyao and Yang, Lijin and Xu, Le and Zhang, Yu and Zhang, Bo and Huang, Wuxiong and Wang, Hesheng},
    title     = {GaussianDWM: 3D Gaussian Driving World Model for Unified Scene Understanding and Multi-Modal Generation},
    booktitle = {Proceedings of the IEEE/CVF Conference on Computer Vision and Pattern Recognition (CVPR)},
    month     = {June},
    year      = {2026},
    pages     = {10656-10667}
}

@article{lin2025depth,
  title={Depth anything 3: Recovering the visual space from any views},
  author={Lin, Haotong and Chen, Sili and Liew, Junhao and Chen, Donny Y and Li, Zhenyu and Shi, Guang and Feng, Jiashi and Kang, Bingyi},
  journal={arXiv preprint arXiv:2511.10647},
  year={2025}
}

@inproceedings{hu2025eva,
  title={EVA-Gaussian: 3D Gaussian-based Real-time Human Novel View Synthesis under Diverse Multi-view Camera Settings},
  author={Hu, Yingdong and Liu, Zhening and Shao, Jiawei and Lin, Zehong and Zhang, Jun},
  booktitle={Proceedings of the IEEE/CVF International Conference on Computer Vision},
  pages={2613--2622},
  year={2025}
}

@ARTICLE{DynAvatar,
  author={Song, Wenfeng and Ye, Zhongyong and Wu, Zhenyu and Li, Shuai and Hou, Xia and Hao, Aimin},
  journal={IEEE Transactions on Visualization and Computer Graphics}, 
  title={DynAvatar: Dynamic 3D Head Avatar Deformation With Expression Guided Gaussian Splatting}, 
  year={2026},
  volume={32},
  number={3},
  pages={2454-2466}
  }
\end{document}